\documentclass{article}
\usepackage{graphicx}

\usepackage[nonatbib,final]{neurips_2021}

\usepackage[square,sort,comma,numbers]{natbib}

\usepackage[utf8]{inputenc} 
\usepackage[T1]{fontenc}    
\usepackage[colorlinks,allcolors=blue!70!black]{hyperref}       
\usepackage{url}            
\usepackage{makecell}
\usepackage{nicefrac}       
\usepackage{microtype}      
\usepackage{bbm}            
\usepackage{multirow}

\usepackage{adjustbox}

\usepackage{caption}
\usepackage{subcaption}

\usepackage{amsmath,amsfonts,amssymb,amsthm,bm}
\usepackage{mathtools}
\usepackage{commath}
\usepackage{float}
\usepackage{algorithm}
\usepackage{algpseudocode}
\usepackage{bbm}

\RequirePackage{enumitem}
\setlist[itemize]{nosep,leftmargin=*,labelwidth=0pt}
\setlist[enumerate]{nosep, leftmargin=*}
\setlist[description]{nosep,leftmargin=.8em}

\RequirePackage[color=yellow!10,textsize=scriptsize,linecolor=gray!80,disable]{todonotes}
\makeatletter
\g@addto@macro{\normalsize}{%
\setlength{\abovedisplayskip}{3pt plus1pt}%
\setlength{\abovedisplayshortskip}{3pt plus1pt}%
\setlength{\belowdisplayskip}{3pt plus1pt}%
\setlength{\belowdisplayshortskip}{3pt plus1pt}}
\let\c@table\c@figure
\makeatother

\RequirePackage{array}
\newcolumntype{?}[1]{!{\vrule width #1}}
\RequirePackage{colortbl}
\RequirePackage{soul}

\newcommand{\vek}[1]{{\mathbf {#1}}}
\newcommand{\vx}{{\vek{x}}}
\newcommand{\vf}{{\vek{f}}}
\newcommand{\va}{{\vek{a}}}

\newcommand{\argmax}{\operatornamewithlimits{argmax}}

\def\mlaas{MLaaS}
\def\match{\text{Agree}}
\newcommand{\service}{S}
\newcommand{\calFull}{Cal:Full}
\newcommand{\calRaw}{Cal:Raw}
\newcommand{\calGold}{Cal:Gold}
\newcommand{\calTemp}{Cal:Temp}
\newcommand{\mx}{$\sigma$}
\def\cX{\mathcal{X}}
\def\cY{\mathcal{Y}}

\def\E{\mathbb{E}}
\def\bmean{{\phi}}
\def\bscale{{\psi}}

\newcommand{\dseed}{{D}}
\def\accgold{\rho}
\def\attspace{\mathcal{A}}
\newcommand{\arms}{$\mathcal{A}$}

\def\attpred{M}
\def\attlist{A}
\def\attrlist{\attlist}
\def\attarm{{\bm{a}}}
\def\accpred{P}
\def\budget{B}

\def\acc{\rho}
\def\iacc{c}

\def\Beta{\mathfrak{B}}
\newcommand{\f}{f}
\newcommand{\g}{g}
\def\fpost{P(\f|\dseed)}
\def\gpost{P(\g|\dseed)}

\def\km{K_1}
\def\ks{K_2}
\def\Emb{\mathcal{V}}

\def\shortname{AAA}
\def\longname{Attributed Accuracy Assay}

\newcommand{\imdbmf}{MF-IMDB}
\newcommand{\celebamf}{MF-CelebA}

\def\PerArmBeta{Beta-I}

\def\BernGP{BernGP}
\def\BetaGP{BetaGP}
\def\BetaAB{\mbox{BetaGP{$\alpha$}{$\beta$}}}
\def\cpred{CPredictor}
\def\DirGP{BetaGP-SL}
\def\DirGPR{BetaGP-SLP}

\def\ztitle{Active Assessment of Prediction Services as Accuracy Surface Over Attribute Combinations}
\title{\ztitle}
\author{%
  Vihari Piratla\thanks{\texttt{vihari@cse.iitb.ac.in}} 
   \qquad
   Soumen Chakrabarty 
   \qquad
   Sunita Sarawagi\\
   Department of Computer Science\\
   Indian Institute of Technology, Bombay
}
\begin{document}
\maketitle

\begin{abstract}
Our goal is to evaluate the accuracy of a black-box classification model, not as a single aggregate on a given test data distribution, but as a surface over a large number of combinations of attributes characterizing multiple test data distributions.  Such attributed accuracy measures become important as machine learning models get deployed as a service, where the training data distribution is hidden from clients, and different clients may be interested in diverse regions of the data distribution. We present \longname{} (\shortname) --- a Gaussian Process (GP)-based probabilistic estimator for such an accuracy surface.
Each attribute combination, called an `arm', is associated with a Beta density from which the service's accuracy is sampled.  We expect the GP to smooth the parameters of the Beta density over related arms to mitigate sparsity.
We show that obvious application of GPs cannot address the challenge of heteroscedastic uncertainty over a huge attribute space that is sparsely and unevenly populated.
In response, we present two enhancements: pooling sparse observations, and regularizing the scale parameter of the Beta densities.
After introducing these innovations, we establish the effectiveness of AAA in terms of both its estimation accuracy and exploration efficiency, through extensive experiments and analysis. Our code and dataset can be found at: \url{https://github.com/vihari/AAA}.
\end{abstract}

\section{Introduction}
\label{sec:Intro}

Increasing concentration of big data and computing resources has resulted in widespread adoption of machine learning as a service (\mlaas).  The best-performing NLP, speech, image and video recognition tools are now provided as network services. \mlaas{} comes with few accuracy specifications or service level agreements, perhaps only leaderboard numbers from benchmarks that may not be closely related to most clients' deployment data distributions. The client, therefore, finds it difficult to choose the best provider without extensive pilot trials~\citep{frugalml20}.  Different clients may need to deploy the service on very different data distributions, with possibly widely different accuracy.  


In such circumstances, we propose that a service provider, or a service standardization agency, publish the accuracy of the classifier, not as one or few aggregate numbers, but as a \emph{surface} defined on a space of input instance \emph{attributes} that capture the variability of consumer expectations.  Indoor/outdoor, day/night, urban/rural may be attributes of input images for visual object recognition tasks. Speaker age, gender, ethnicity/accent may be attributes of input audio for speech recognition tasks.  We call a combination of attributes in their Cartesian space an \emph{arm} (borrowing from bandit terminology)\footnote{\figurename~\ref{fig:task:stats1c} shows an example of diverse accuracy over arms.}.  
The labeled instances used by the service provider may not represent or cover well the space of attributes of interest to subscribers. Labeled data may be proprietary and inaccessible to prospective consumers and standardization agencies. Whoever estimates the accuracy surface, therefore, needs to \emph{actively} select instances from an unlabeled pool for labeling, presumably within a restricted budget, to adequately cover the attribute space.

Several recent studies have highlighted the variability in accuracy across data sub-populations~\citep{Subbaswamy21eval,Sagawa19}, specifically in the context of fairness \citep{gendershades,modelcards19,JiSS20neurips}, and also proposed active estimation techniques of sub-population accuracy~\citep{JiLR20,Miller21}.  We solve a more general problem where the space of arms (sub-population) defined by the Cartesian space of attributes grows combinatorially.
This inevitably leads to extreme sparsity of labeled instances for many arms.
A central challenge is how to smooth the estimate across related arms while faithfully representing the uncertainty for active exploration.

We present \longname{} (\shortname) --- a practical system that estimates accuracy, together with the uncertainty of the estimate, as a function of the attribute space.  \shortname{} uses these estimates to drive the sampling policy for each attribute combination. Gaussian Process (GP) regression is a natural choice to obtain smooth probabilistic accuracy estimates over arm attributes. However, a straightforward GP model fails to address the challenge of heteroscedasticity that we face with uneven and sparse supervision across arms.  We model arm-specific service accuracy as drawn from a Beta density that is characterized by mean and scale parameters, which are sampled from two GPs that are informed by suitable trained kernels over the attribute space.  We propose two further enhancements to the training of this model.
%
First, we recognize an over-smoothing problem with GP's estimation of the Beta scale parameters, and propose  
a Dirichlet likelihood to supervise the relative values of scale across arms. Second, we recognize that arms with very low support interfere with learning the kernel parameters of the GPs.  We mitigate this by pooling observations across related arms.
With these fixes, \shortname{} achieves the best estimation performance among competitive alternatives.

Another practical challenge in our setting is that some attributes of instances are not known exactly. For example, attributes, such as camera shutter speed or speaker gender, may be explicitly provided as meta information attached with instances.  
But other attributes, such as indoor/outdoor, or speaker age, may have to be estimated noisily via another (attribute) classifier, because accurate human-based acquisition of attributes would be burdensome.   \shortname{} also tackles uncertain attribute inference.  Its attribute classifiers are trained on a small amount of labeled data and their error rates are modeled in a probabilistic framework.

We report on extensive experiments using several real data sets. Comparison with several estimators based on Bernoulli arm parameters, Beta densities per arm, and even simpler forms of GPs on the arm Beta distributions, shows that \shortname{} is superior at quickly cutting down arm accuracy uncertainty.

Summarizing, our contributions are:
\begin{itemize}
\item We motivate and define the problem of accuracy surface  estimation over a large space of attribute combinations.  

\item Our proposed estimator \shortname\ fits a Beta density for every attribute combination (arm), with its parameters smoothed via two GPs to capture heteroscedastic uncertainty of each arm's accuracy under limited data settings. 

\item We propose two important components included in \shortname: 1)~a Dirichlet regularization to control over-smoothing of the Beta scale parameters, and 2)~pooled observations to reduce over-fitting of a GP-associated kernel to sparse arms.

\item We show significant gains in terms of  both estimation quality and the efficiency of exploration on four real classification models  compared to existing methods. \shortname{} obtains an average 80\% reduction in macro averaged square error over the existing methods. 
\end{itemize}

\section{Problem Setup}
\label{sec:problem}

Our goal is to evaluate a given machine learning service model $S$ used by a diverse set of consumers. 
The service $S:\cX\mapsto\cY$ could be any predictive model that, for an input instance $\vx \in \cX$, assigns an output label $\hat{y} \in \cY$, where $\cY$ is a discrete label space. Let $y(\vx)$ denote the true label of $\vx$ and $\match(y,\hat{y})$ denote the match between the two labels.  For scalar classification, $\match(y,\hat{y})$ is in \{0,1\}. For structured outputs, e.g., sequences, we could use measures like BLEU scores in [0,1].   Classifiers are routinely evaluated on their expected accuracy on a data distribution $P(\cX,\cY)$:
\begin{align}
\accgold = \E_{P(\vx,y)}[\match(y, S(\vx)] \label{eq:OneGamma}
\end{align}
We propose to go beyond this single measure and define accuracy as a surface over a space of attributes of the input instances.  
Let $\attrlist$ denote a list of $K$ attributes that capture the variability of consumer expectation on which the service $S$ will be deployed.  For instance, visual object recognition is affected by the background scene, and facial recognition is affected by demographic attributes.  We use $\attrlist(\vx) \in \attspace$ to denote the vector of values of attributes of input $\vx$ and $\attspace$ to denote the Cartesian product of the domains of all attributes.  An attribute could be discrete, e.g., the ethnicity of a speaker; Boolean, e.g., whether a scene is outdoors/indoors; or continuous, e.g., the age of the speaker in speech recognition. Some of the attributes of $\vx$, for example the camera settings of an image, may be known exactly, and others may only be available as a distribution $\attpred_k(a_k|\vx)$ for an attribute $a_k \in \attrlist$, obtained from a pre-trained probabilistic classifier.

Generalizing from a single global expected accuracy \eqref{eq:OneGamma}, we define the accuracy surface  $\accgold:\attspace \rightarrow [0,1]$ of a service $S$ at each attribute combination $\attarm \in \attspace$, given a data distribution $P(\cX,\cY)$, as
\begin{equation}
\accgold(\attarm) = \E_{P(\vx,y|A(\vx)=\attarm)}[\match(y, S(\vx)]
\end{equation}
Our goal is to provide an estimate of $\accgold(\attarm)$ given two kind of  data sampled from $P(\cX,\cY)$: a 
small labeled sample $\dseed$, and a large unlabeled sample~$U$.  
In addition, we are given a budget of $\budget$ instances for which we can seek labels $y$ from a human by selecting them from~$U$.  Applying $M_k$ to all of $U$ is, however, free of cost.

We aim to design a probabilistic estimator for~$\accgold(\attarm)$, which we denote as $\accpred(\acc|\attarm)$ where $\acc \in [0,1]$ and $\attarm \in \attspace$.  This is distinct from active learning, which selects instances to train the learner toward greater accuracy, and also active accuracy estimation~\citep{JiLR20}, which does not involve a surface over~$\attarm$s.  We also show that standard tools to regress from $\attarm$ to $\accgold$ are worse than our proposal.

We measure the quality of our estimate as the square error between the gold accuracy $\accgold(\attarm)$ and the mean of the estimated accuracy distribution $\accpred(\acc|\attarm)$.  Our estimator distribution naturally gives an idea of the posterior variance of accuracy estimate of each attribute combination, which we use for uncertainty-based exploration.  

\section{Proposed Estimator}
\label{sec:aaa}

We will first review recent work that leads to candidate solutions to our problem, discuss their limitations, and finally present our solution.
Initially, to keep the treatment simple, we assume $\attlist(\vx)$ and gold $y$ (hence $c=\match(S(x), y)$, the service correctness bit)
is known for all instances.  Later in this section, we remove these assumptions.

The simplest option is to ignore any relationship between arms, and, for each arm $\attarm$, fit a suitable density over $\accgold(\attarm)$.  When this density is sampled, we get a number in $[0,1]$, which is like a coin head probability used to sample correctness bits~$c$.
For representing uncertainty of accuracy values (which are ratios between two counts), the \href{https://en.wikipedia.org/wiki/Beta_distribution}{Beta distribution} $\Beta(\cdot,\cdot)$ is a natural choice.  
We call this baseline method \textbf{\PerArmBeta}.

The variance of the estimated Beta density can be used for actively sampling arms.  \citet{JiLR20} describe a related scenario, stressing on active sampling.  However, this approach cannot share observations or smooth the estimated density at a sparsely-populated arm with information from similar arms.  In our real-life scenario, we expect accuracy surface smoother and the number of arms to be large enough that many arms will get very few, if any, instances.

The second baseline method, which we call \textbf{\BernGP}, is to view the $(\attarm,\iacc)$ instances in $\dseed$ as a standard classification data set with the binary $\iacc$ values as class label and $\attarm$ as input features.  Given the limited data, we can use the well-known GP classification approach \citep{HensmanMG15} for fitting smooth values $\acc$ as a function of~$\attarm$. %
Suppose the arms $\attarm$ can be embedded to $\Emb(\attarm)$ in a suitable space induced by some similarity kernel.  In this embedding space, we expect the accuracy of $S$ to vary smoothly.
Given a kernel $K_1(\attarm, \attarm')$ to guide the extent of sharing of information across arms, a standard form of this GP would be
\begin{align}
P(\iacc|\attarm) &= 
\text{Bernoulli}(\iacc; \operatorname{sigmoid}(f_\attarm));
\quad f \sim GP(0, K_1).
\label{eq:bernLL}
\end{align}
The GP can give estimates of uncertainty of $\accgold(\attarm)$, which may be used for active sampling of arms.

As we will demonstrate, such GP-imposed estimate of uncertainty of $\accgold(\attarm)$ is inadequate, because it loses sight of the number of supporting observations at each arm, which could be very diverse.
This is because the standard GP assumption of homoscedasticity, that is, identical noise around each arm is violated when observations per arm differ significantly.  We therefore need a mechanism to separately account for the uncertainty at each arm, even the unexplored ones, to guide the strategy for actively collecting more labeled data. 


\subsection{The basic \BetaGP{} proposal}
\label{sec:betagp}
We model arm-specific noise by allowing each arm to represent the uncertainty of $\rho_a$, not just by an underlying GP as in \BernGP\ above, but also by a separate scale parameter.  Further, the scale parameter is smoothed over neighboring arms using another GP. The influence of this scale on the uncertainty of $\rho_a$ is expressed by a Beta distribution as follows: 
\begin{align}
\accpred(\acc|\attarm) &\sim \Beta(\acc; \bmean(\f_\attarm), \bscale(\g_\attarm)) \label{eq:model}\\
\bmean(\f_\attarm) &= \operatorname{sigmoid}(\f_\attarm),
\qquad \f \sim GP(0, \km),  \label{eq:priorm}  \\
\bscale(\g_\attarm) &= \log(1+e^{\g_\attarm}),
\qquad \g \sim GP(0, \ks), \label{eq:priors}
\end{align}
where we use $\bmean(\bullet), \bscale(\bullet)$ to denote the parameters of the Beta distribution at arm $\attarm$.
The Beta distribution is commonly represented via $\alpha,\beta$ parameters whereas we chose the less popular mean ($\bmean$) and scale ($\bscale$) parameters.  While these two forms are functionally equivalent with $\bmean = \frac{\alpha}{\alpha+\beta}, \bscale=\alpha+\beta$, we preferred the second form because imposing GP smoothness across arms on the mean accuracy and scale seemed more meaningful.  We validate this empirically in the Appendix~\ref{sec:appendix:betaab}.  

Two kernel functions $\km(\attarm,\attarm')$, $\ks(\attarm,\attarm')$ defined over pairs of
arms $\attarm, \attarm' \in \attspace$ 
control the degree of smoothness among the Beta parameters across the arms.  We use an RBF kernel defined over learned shared embeddings $\Emb(\attarm)$:
\begin{align}
    \km(\attarm, \attarm') = s_1\exp\left[-\tfrac{\|\Emb(\attarm) - \Emb(\attarm')\|^2}{l_1}\right],
    \qquad
     \ks(\attarm, \attarm') = s_2\exp\left[-\tfrac{\|\Emb(\attarm) - \Emb(\attarm')\|^2}{l_2}\right]
    \label{eq:kernel}
\end{align}
where $s_1,s_2,l_1,l_2$ denote the scale and length parameters of the two kernels.  The scale and length parameters are learned along with the parameters of embeddings $\Emb(\attarm)$ during training.

Initially, we assume we are given a labeled dataset $\dseed = \{(\vx_i, \attarm_i, y_i):i=1\ldots,I\}$ with attribute information available.  Using predictions from the classification service $\service$, we associate a 0/1 accuracy $\iacc_i=\match(y_i, \service(\vx_i))$.  We can thus extend $\dseed$ to $\{(\vx_i, \attarm_i, y_i, c_i): i\in[I]\}$.

Let $\iacc_\attarm=\sum_{i: A(\vx_i)=\attarm} \iacc_i$ denote the total agree score in arm $\attarm$.  Let $n_\attarm$ denote the total number of labeled examples in arm~$\attarm$. 
The likelihood of all observations given functions $\f,\g$ decomposes as a product of Beta-binomial\footnote{The ${n_a \choose c_a}$ term does not apply since we are given not just counts but accuracy $c_i$ of individual points.} distributions at each arm as follows:
\begin{align}
\Pr(\dseed|\f,\g) &= \prod_\attarm 
\int_\rho \rho^{\iacc_a}(1-\rho)^{n_\attarm-\iacc_\attarm}\, \Beta(\rho|\bmean(\f_{\attarm}), \bscale(\g_{\attarm}))) \text{d}\rho.
\label{eq:pointLL} \\
&=  \prod_\attarm  \frac{\text{B}(\bmean(\f_{\attarm})\bscale(\g_\attarm) + \iacc_a,   (1-\bmean(\f_{\attarm}))\bscale(\g_\attarm)  + n_\attarm -\iacc_a) }{
\text{B}(\bmean(\f_{\attarm})\bscale(\g_\attarm),   (1-\bmean(\f_{\attarm}))\bscale(\g_\attarm))
},
\label{eq:pointLLSimple}
\end{align}
where \text{B} is the Beta function, and the second expression is a rewrite of the \href{https://en.wikipedia.org/wiki/Beta-binomial_distribution}{Beta-binomial likelihood}.

During training we calculate the posterior distribution of functions $f,g$ using the above data likelihood $\Pr(D|f,g)$ and GP priors given in eqns.~\eqref{eq:priorm} and~\eqref{eq:priors}.  The posterior cannot be computed analytically given our likelihood, so we use variational methods. Further, we reduce the $\mathcal{O}(|\attspace|^3)$ complexity of posterior computation, using the inducing point method of \citet{HensmanMG15}, whereby we learn $m$ locations $\mathbf{u} \in \mathbb{R}^{d\times m}$, mean $\mu \in \mathbb{R}^m$, and covariance $\Sigma \in \mathbb{R}^{m\times m}$ of inducing points. Doing so brings down the complexity to $\mathcal{O}(m^2|\attspace|), m \ll |\attspace|$. These parameters are learned end to end with the parameters of the neural network used to extract embeddings $\Emb(\attarm)$ of arms $\attarm$, and kernel parameters $s_1,s_2,\ell_1,\ell_2$. 
%
We used off-the-shelf Gaussian process library: GPyTorch~\citep{gpytorch} to train the above likelihood with variational methods.
Details of this procedure can be found in the Appendix~\ref{sec:appendix:gp}. We denote the posterior functions as $\fpost, \gpost$. 
Thereafter, the mean estimated accuracy for an arm $\attarm$ is computed as
\begin{equation}
\label{eq:our:est}
\E(\rho|\attarm) = \E_{\f\sim\fpost} [\bmean(\f_\attarm)].
\end{equation}
We call this setup \textbf{\BetaGP}.
Next, we will argue why \BetaGP{} still has serious limitations, and offer mitigation measures.




\subsection{Supervision for scale parameters}
\label{sec:aaa:sl}

We had introduced the second GP $g_\attarm$ to model arm-specific noise, and similar techniques have been proposed earlier by \citet{LzaroGredilla2011VariationalHG, Kersting2007MostLH,Goldberg1997}, but for heteroscedasticity in Gaussian observations.  
However, we found the posterior distribution of scale values $\psi(g_\attarm)$ at each arm tended to converge to similar values, even across arms with orders of magnitude difference in number of observations $n_\attarm$.  On hindsight, that was to be expected, because the data likelihood~\eqref{eq:pointLL} increases monotonically with scale $\psi_\attarm$. The only control over its converging to $\infty$ is the GP prior $g \sim GP(0,K_2)$. 
In Appendix~\ref{sec:appendix:synth}, we illustrate this phenomenon with an example. 
%
%
We propose a simple fix to the scale supervision problem. We expect the relative values of scale across arms to reflect the distribution of the proportion of observations $\frac{n_a}{n}$ across arms (with $n=\sum_\attarm n_\attarm$).  We impose a joint Dirichlet distribution using the scale of arms $\psi(g_\attarm)$ as parameters, and write the likelihood of the observed proportions as (with $\Gamma$ denoting \href{https://en.wikipedia.org/wiki/Gamma_function}{Gamma function}):
\begin{equation}
\log\Pr(\{n_\attarm\}|g) = \sum_\attarm ((\psi(g_\attarm)-1) \log \frac{n_a}{n} - \log \Gamma(\psi(g_\attarm)) + \log\Gamma(\textstyle\sum_\attarm \psi(g_\attarm))
\end{equation}
We call this \textbf{\DirGP}. With this as an additional term in the data likelihood, we obtained significantly improved uncertainty estimates at each arm, as we will show in the experiment section. 


\subsection{Pooling for sparse observations}
\label{sec:aaa:pool}

Recall that the observations are accumulation of 1/0 agreement scores for all instances that belong to an arm. Given the nature of our problem, arms have varying levels of supervision, and also highly varying true accuracy values. 
Even when the available labeled data is large, many arms will continue to have sparse supervision because they represent rare attribute combinations.
%
The combination of high variance observations and sparse supervision could lead to learning of non-smooth kernel parameters. 
In Appendix~\ref{sec:appendix:synth}, we demonstrate with a simple setting that GP parameters learned on noisy observations under-represent the smoothness of the surface.
The situation is further aggravated when learning a deep kernel. This problem has resemblance to ``collapsing variance problem''~\citep{murphyml} such as when Gaussian mixture models overfit on outliers or when topic models overfit a noisy document in the corpus.
Instead of depending purely on GP priors to smooth over these noisy observations, we found it helpful to also externally smooth noisy observations.  For each arm $\attarm$ with observations below a threshold, we mean-pool observations from some number of nearest neighbors, weighted by their kernel similarity with~$\attarm$.  We will see that such external smoothing resulted in significantly more accurate estimates particularly for arms with extreme accuracy values.  We call this method \textbf{\DirGPR} (note that this also includes the scale supervision objective described in the previous section).
Two other mechanisms take us to the full form of the \textbf{\shortname} system, which we describe next.

\subsection{Exploration}
\label{sec:aaa:Explore}

The variance estimate of an arm informs its uncertainty and is commonly used for efficient exploration~\citep{Schulz18tutorial}. 
Let $\fpost,\gpost$ denote the learned posterior distribution of the GPs.  Using these, the estimated variance at an arm is given as:
\begin{align}
\mathbb{V}(\rho|\attarm) = \E_{\f\sim\fpost,\g\sim\gpost} 
\left[ \textstyle \int_\rho (\rho-\E(\rho|\attarm))^2
      \Beta(\rho; \bmean(\f_\attarm ), \bscale(\g_\attarm ))
      \text{d}\rho \right]
\label{eqn:mean_variance}
\end{align}
where the expected value is given in eqn.~\eqref{eq:our:est}.  We use sampling to estimate the above expectation.  The arm to be sampled next is chosen as the one with the highest variance among unexplored arms. We then sample an unexplored example with highest affiliation ($P(\va\mid\vx)$) with the chosen arm. 





\subsection{Modeling Attribute Uncertainty}
\label{sec:aaa:AttribNoise}
Recall that attributes of an instance $\vx$ are obtained from  models $M_k(a_k|\vx),~~k \in [K]$, which may be highly noisy for some attributes. Thus, we cannot assume a fixed attribute vector $A(\vx)$ for an instance $\vx$.  We address this by designing a model that can combine these noisy estimates into a joint distribution $P(\va|\vx)$ using which, we can fractionally assign each instance $\vx_i$ across arms.   A baseline model for $P(\va|\vx)$ would be just the product $\prod_{k=1}^K M_k(a_k|\vx)$. However, 
we expect values of attributes to be correlated (e.g. attribute `high-pitch' is likely to be correlated with gender `female'). Also, the probabilities  $M_k(a_k|\vx)$ may not be well-calibrated.  

We therefore propose an alternative joint model that can both recalibrate individual classifiers via temperature scaling~\cite{GuoPSW17}, and model their correlation.   We have a small seed labeled dataset $\dseed$ with gold attribute labels, independent noisy distributions from each attribute model $M_k(a_k|\vx)$, and an unlabeled dataset $U$. We prefer simple factorized models.
%
%
We factorize $\log\Pr(\va|\vx)$ as a sum of temperature-weighted logits and a joint (log) potential as shown in expression~\eqref{eqn:factor:assume} below. 

\begin{align}
\log \Pr(\va|\vx) = \log\Pr(a_1, a_2, \cdots ,a_K|\vx) = \sum_{k=1}^K t_k\log M_k(a_k|\vx) +
N(a_1, a_2, \cdots, a_K) \label{eqn:factor:assume}
\end{align}
Here $N$ denotes a dense network to model 
the correlation between attributes, and  $t_1,\ldots,t_K$ denote the temperature parameters used to rescale noisy attribute probabilities.
%
The maximum likelihood over $\dseed$ is
$\max_{t,N} \sum_{(\vx_i,\attarm_i) \in \dseed} \log \Pr(\va_i|\vx_i)$ 
%
%
\begin{align}
=\max_{t,N} \sum_{(\vx_i, \attarm_i) \in \dseed} \big\{ \textstyle
\sum_{k=1}^K t_k \log M_{k}(a_{ik}|\vx_i)+N(a_{i1}, \ldots a_{iK}) - \log (Z_i) \big\}    
  \label{eqn:max_likelihood}
\end{align}
$Z_i$ denotes the partition function for an example $\vx_i$ which requires summation over~$\attspace$.  Exact computation of $Z_i$ could be intractable especially when \arms\ is large. In such cases, $Z_i$ can be approximated by sampling. In our case, we could get exact estimates.

In addition to $\dseed$, we use the unlabeled instances~$U$ with predictions from attribute predictors filling the role of gold-attributes. Details on how we train the parameters on large but noisy $U$ and small but correct $\dseed$ can be found in the Appendix~\ref{sec:appendix:calibration}.

The estimation method of \DirGPR{} with variance based exploration and calibration described here constitute our proposed estimator: \shortname{}. Detailed pseudo-code of \shortname{} is given in the Appendix~\ref{sec:appendix:pseudocode}.

\section{Experiments}
\label{sec:Expt}

Our exploration of various methods and data sets is guided by the following research questions.
\begin{itemize}
\item How do various methods for arm accuracy estimation compare?
\item To what extent do \BetaGP, scale supervision and pooled observations help beyond \BernGP?
\item For the best techniques from above, how 
do various active exploration strategies compare?
\item How well does our proposed model of attribute uncertainty work?
\end{itemize}

\subsection{Data sets and tasks}
\label{sec:expt:dataset}

We experiment with two real data sets and tasks.  Our two tasks are male-female gender classification with two classes and animal classification with 10 classes.

\paragraph{Male-Female classification (MF):} 
CelebA \citep{CelebA} is a popular celebrity faces and attribute data set that identifies the gender of celebrities among 39 other binary attributes.  The label is gender.  The accuracy surface spans various demographic, style, and personality related attributes.  We hand-pick a subset of 12 attributes that include attributes that we deem important for gender classification among some other gender-neural attributes such as if the subject is young or wearing glasses (see Appendix~\ref{sec:appendix:task} for more details). We used a random subset of 50,000 examples from the dataset for training classifiers on each of the 12 attributes using a pretrained ResNet-50 model. The remaining 150,000 examples in the data set are set as the unlabeled pool from which we actively explore new examples for human feedback. The twelve binary attributes make up for $2^{12}=4,096$ attribute combinations.

\paragraph{Animal classification (AC):} 
COCO-Stuff \citep{COCOS} provides an image collection.  For each image, labels for foreground (cow, camel) and background (sky, snow, water) `stuff' are available.  Visual recognition models often correlate the background scene with the animal label such as camel with deserts and cow with meadows.  Thus, foreground labels are our regular $y$-labels while background stuff labels supply our notion of attributes.

We collapse fine background labels into five coarse labels using the dataset provided label hierarchy. These are: water, ground, sky, structure, furniture.  The Coco dataset has around 90 object (foreground) labels.  Here we use a subset of 10 labels corresponding to animals. We take special care to filter out images with multiple/no animals and adapt the pixel segmentation/classification task to object classification (see the Appendix~\ref{sec:appendix:task} for more details). The image is further annotated with the five binary labels corresponding to five coarse stuff labels. The scene descriptive five binary labels and ten object labels make up for $32{\times}10=320$ attribute combinations.  


\subsection{Service Models}
\label{sec:expt:services}

For the MF task, we use two service models~($S$).
\textbf{MF-CelebA} is a service model for gender classification.  To simulate separate $\dseed$ and $U$, it is trained on a random subset of CelebA with a ResNet50 model.  
\textbf{MF-IMDB} is a publicly available\footnote{\url{https://github.com/yu4u/age-gender-estimation}} classifier trained on IMBD-Wiki dataset, also using the ResNet50 architecture.  The attribute predictors are trained using ResNet50 on a subset of the CelebA dataset for both service models.

\begin{figure}
\begin{minipage}[t]{0.3\textwidth}
   \vspace{0pt}
    \centering
    \includegraphics[width=\linewidth]{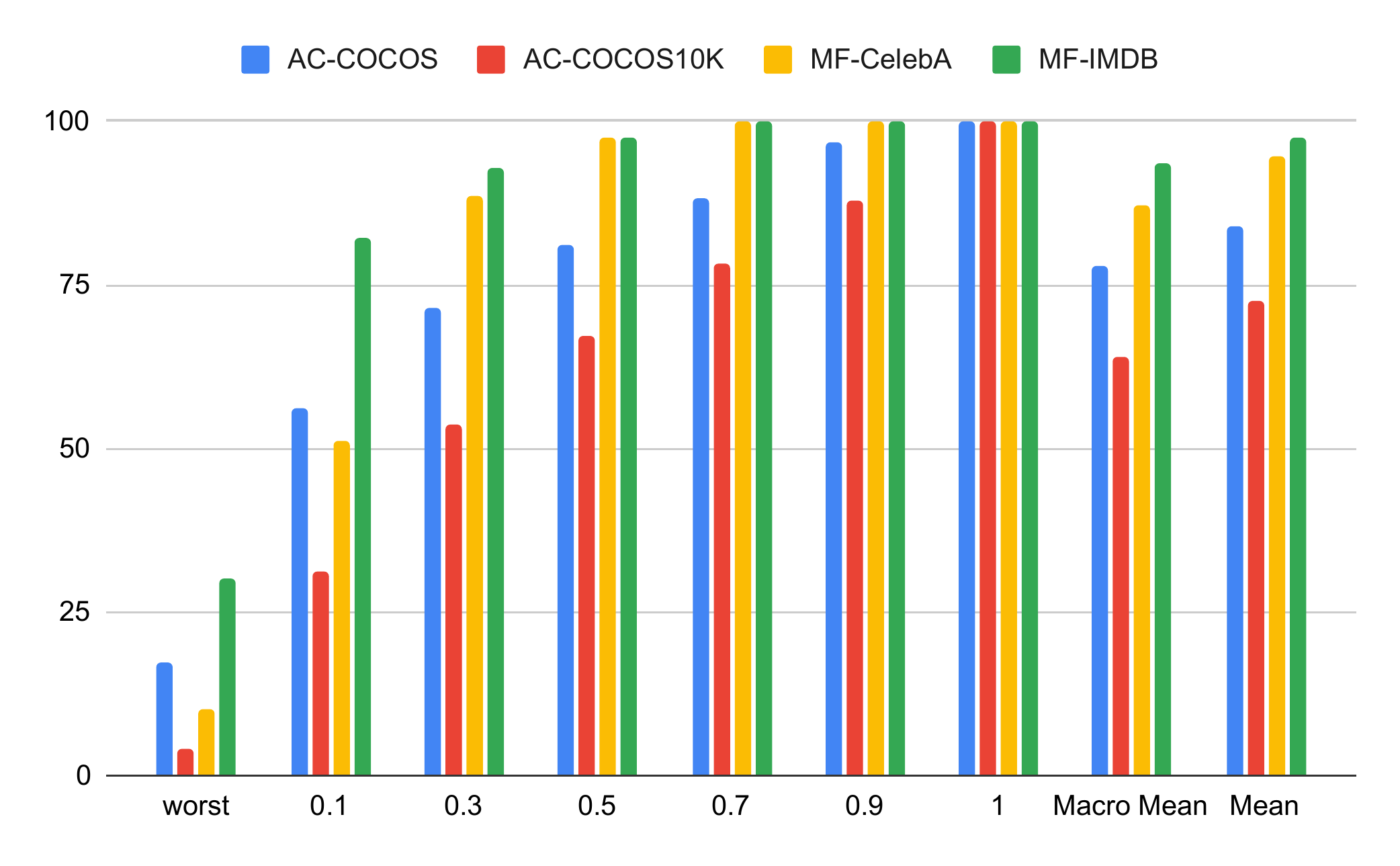}
    \captionof{figure}{Macro and micro averaged accuracy (right most) and ten quantiles (x-axis) of per-arm accuracy (y-axis).}
    \label{fig:task:stats1c}
\end{minipage}\hfill
\begin{minipage}[t]{0.68\textwidth}
\vspace{0pt}
\centering
\resizebox{0.95\hsize}{!}{%
\tabcolsep 2pt
\begin{tabular}{|l|c|c|c|c|} \hline
Service$\rightarrow$ & {\bf AC-COCOS10K} & {\bf AC-COCOS} & {\bf MF-IMDB} & {\bf MF-CelebA}\\ \hline
\cpred & 5.4 / 15.0 & 3.2 / 9.4 & 1.2 / 8.2 & 5.2 / 35.9 \\
\PerArmBeta & 7.0 / 15.6 & 4.3 / 10.0 & 1.6 / 8.4 & 4.7 / 30.3\\
\BernGP & 7.0 / 13.2 & 3.5 / 8.6 & 1.7 / 7.6 & 4.9 / 28.1\\
\BetaGP & 7.1 / 14.3 & 3.3 / 7.9 & 2.2 / 6.6 & 4.6 / 25.9\\
\DirGP & 5.3 / 11.7 & 2.8 / 6.8 & 1.4 / 4.4 & 4.1 / 22.6 \\
\rowcolor{green!10}
\DirGPR & 4.7 / 10.4 & 2.8 / 5.7 & 1.4 / 3.9 & 4.3 / 23.3\\\hline
 \end{tabular}
}
\captionof{table}{Comparing different estimation methods on labeled data  size 2000 across four tasks. No exploration is involved.  Each cell shows two numbers in the format ``macro MSE / worst MSE'' obtained over three runs.  \DirGPR{} generally gives the lowest MSE.}
\label{tab:est}
\end{minipage}
\end{figure}

For the AC task, we use two publicly available\footnote{\url{https://github.com/kazuto1011/deeplab-pytorch/}} service models~($S$).
\textbf{AC-COCOS} was trained on COCOS data set with 164K examples.
\textbf{AC-COCOS10k} was trained on COCOS10K, an earlier version of COCOS with only 10K instances. We use these architectures for both label and attribute prediction.
%
See Appendix~\ref{sec:appendix:task},~\ref{sec:appendix:surface_stats} for more details on accuracy surface, attribute predictor, service models and their architecture. 
In Figure~\ref{fig:task:stats1c}, we illustrate some statistics of the shape of the accuracy surface for the four dataset-task combinations. Although $S$'s mean accuracy (rightmost bars) is reasonably high, the accuracy of the arms in the 10\% quantile is abysmally low, while arms in the top quantiles have near perfect accuracy. This further motivates the need for an accuracy surface instead of single accuracy estimate.

\subsection{Methods Compared}

We compare the proposed estimation method \shortname{} against natural baselines, alternatives, and ablations.
Some of the methods, such as \textbf{\PerArmBeta}, \textbf{\BernGP} and \textbf{\BetaGP}, we have already defined in Section~\ref{sec:aaa}.  
We train methods \BernGP{} and \BetaGP{} using the default arm-level likelihood.  
We also separately evaluate the impact of our fixes on \BetaGP{} with only scale supervision:~{\bf \DirGP{}} and along with mean pooling:~{\bf \DirGPR{}}.
We also include a trivial baseline: {\bf \cpred{}} which fits all the arms with a global accuracy estimated using gold $\dseed$. We do not try sparse observation pooling with \PerArmBeta{} since there is no notion of per-arm closeness. We also skip it on \BernGP{} since it is worse than \BetaGP{} as we will show below. Recall that {\PerArmBeta} modeling is related to~\citet{JiLR20}.


\subsection{Other experimental settings}
\label{sec:expt:other}
\vspace{-0.2cm}
\textbf{Gold accuracies~$\accgold(\attarm)$:}
We compute the oracular accuracy per arm using the gold attribute/label values of examples in $U$ which we treat as unlabeled during exploration. For every arm with at least five examples, we set its accuracy to be the empirical estimate obtained through the average correctness of all the examples that belong to the arm. We discard and not evaluate on any arms with fewer than five examples since their true accuracy cannot reliably be estimated. 

\textbf{Warm start:} 
We start with 500 examples having gold attributes+labels to warm start all our experiments. The random seed also picks this random subset of 500 labeled examples. We calculate the overall accuracy of the classifier on these warm start examples as  $\hat{\accgold} = ({\sum_i c_i})/({\sum_i 1})$.
For all arms we warm start their observation with $c_\attarm = \lambda\hat{\accgold}, n_\attarm=\lambda$
where $\lambda=0.1$, a randomly picked low value.


Unless otherwise specified, we give equal importance to each arm and report MSE macroaveraged over all arms. Along with macro MSE, we also sometimes report MSE on the subset of 50 worst (true-)accuracy arms, referred to as worst MSE.  
We report other aggregate errors in the Appendix~\ref{sec:appendix:metrics}.
All the numbers reported here are averaged over three runs each with different random seed. The initial set of warm-start examples ($\dseed$) is also changed between the runs. In the case of \DirGPR{}, for any arm with observation count below 5, we mean pool from its three closest neighbours. 


In the following Sections:~\ref{sec:expt:est} and~\ref{sec:expt:exp}, we compare various estimation and exploration strategies with $P(\va{\mid}\vx)$ noise calibrated as described in Section~\ref{sec:aaa:AttribNoise}. In Section~\ref{sec:expt:calib}, we study different forms of calibration and demonstrate the superiority of our proposed calibration technique of Equation~\eqref{eqn:factor:assume}.

\begin{figure*}[t]
\centering
\begin{tabular}{cccc} \tabcolsep 1pt
\includegraphics[width=.22\hsize]%
{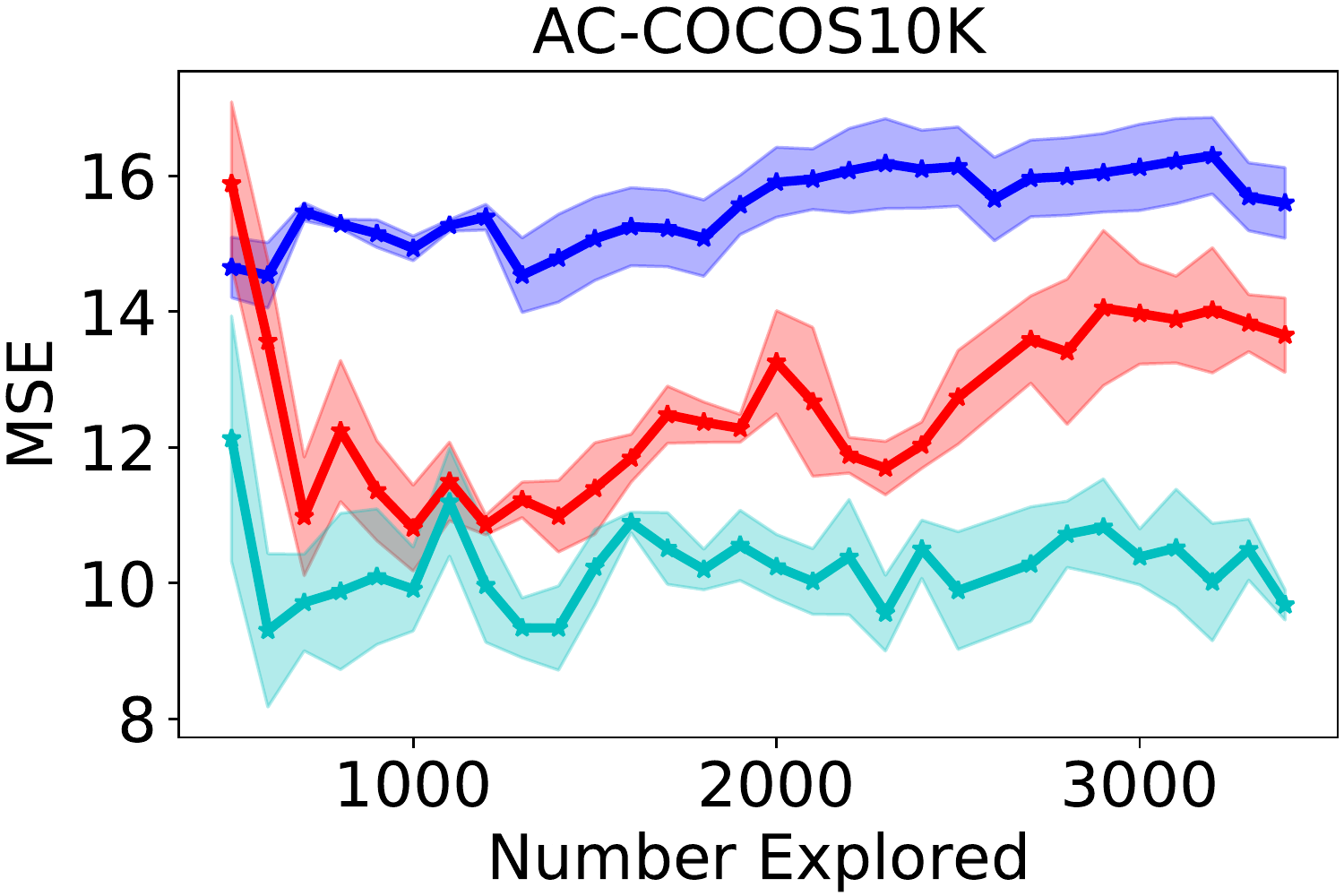} &
\includegraphics[width=.22\hsize]%
{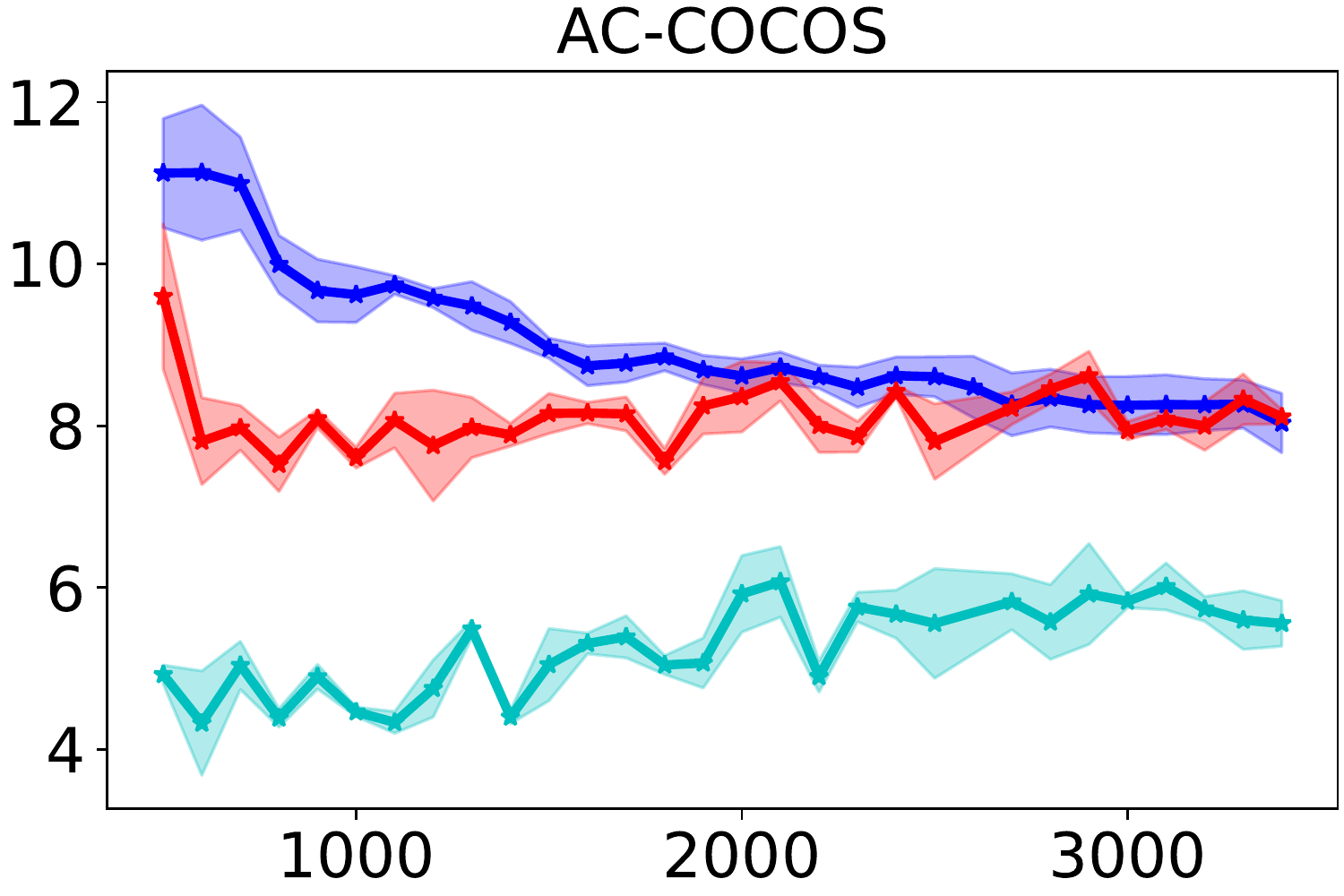} &
\includegraphics[width=.22\hsize]%
{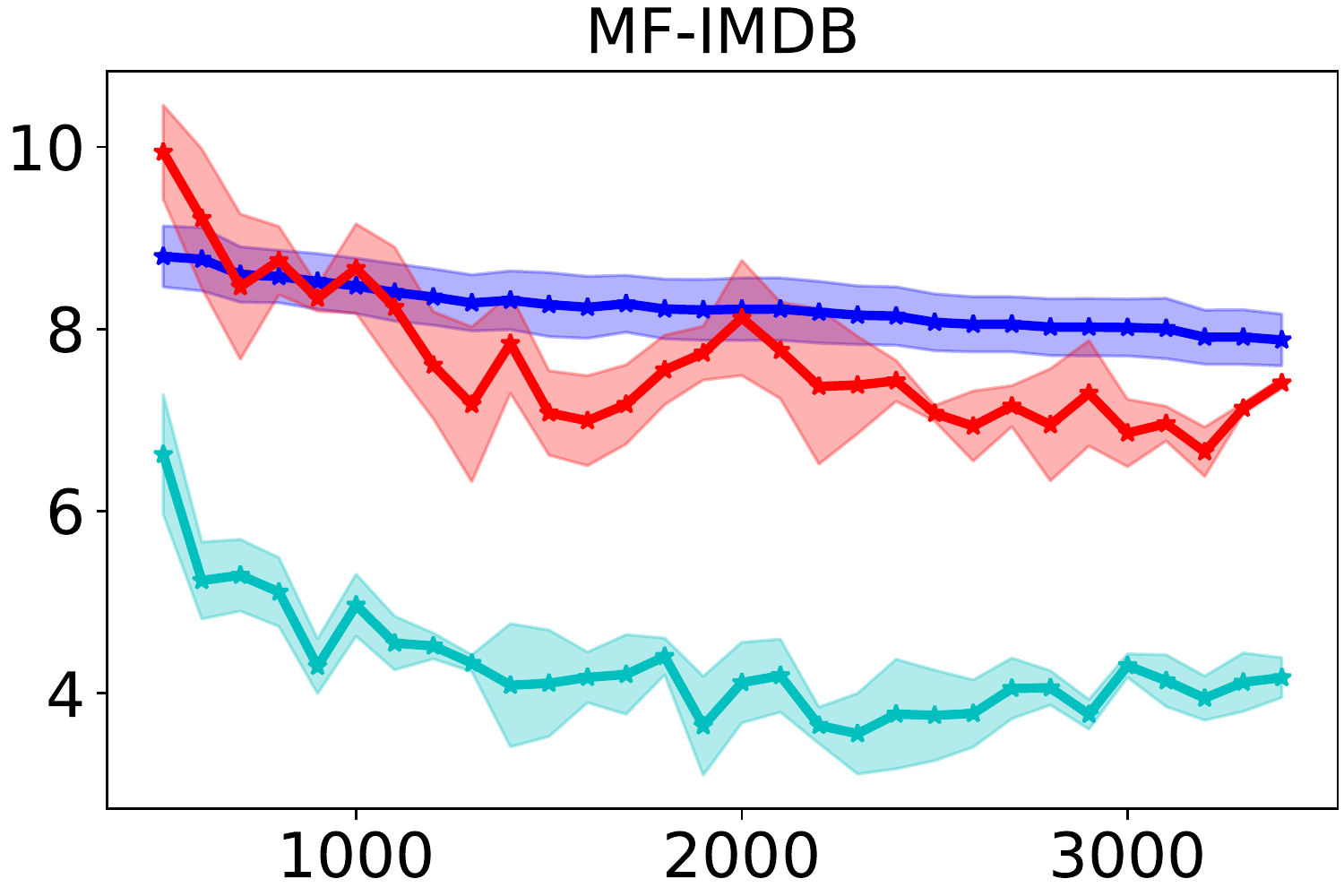} &
\includegraphics[width=.22\hsize]%
{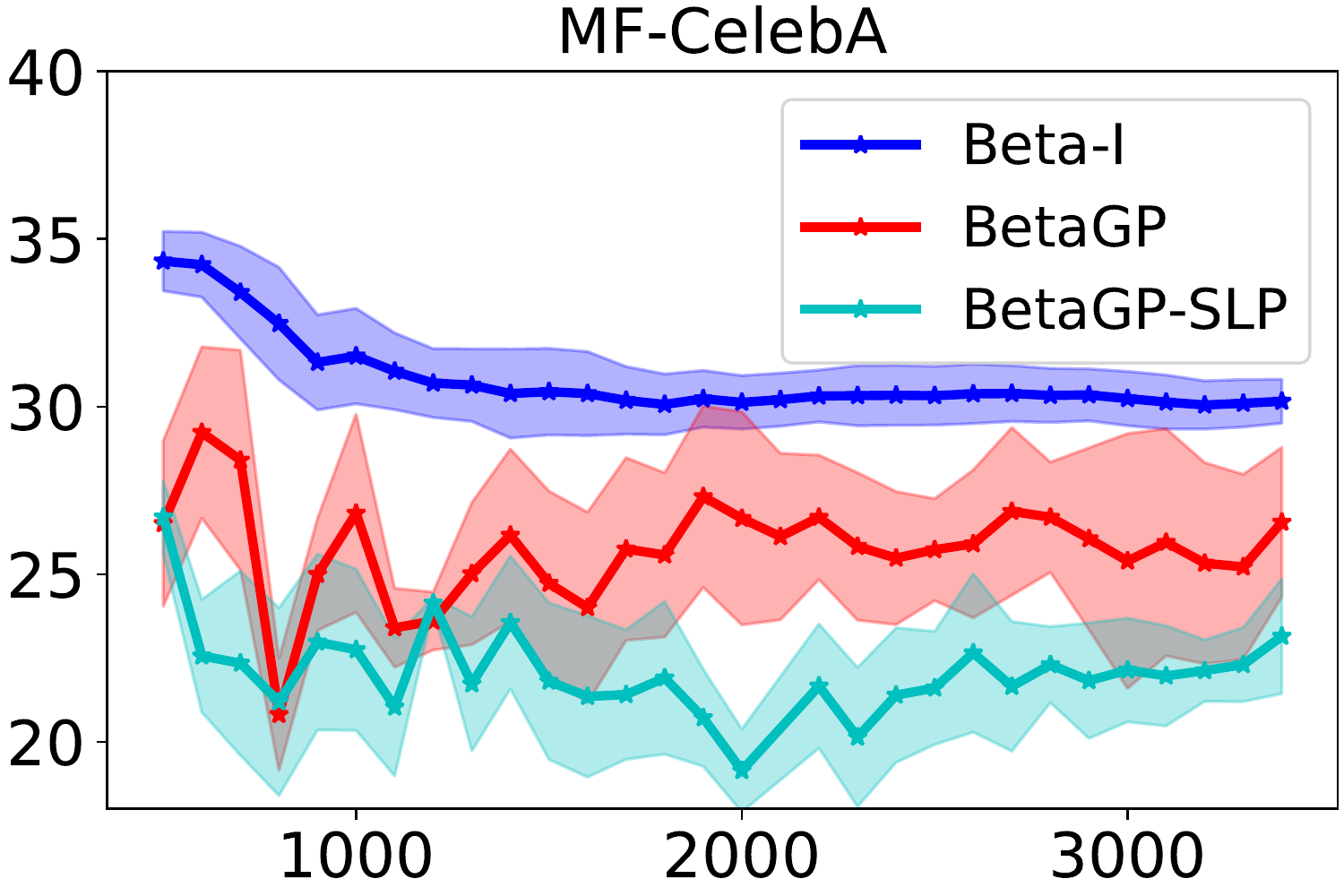}
\end{tabular}
\caption{Comparison of estimation methods using worst MSE metric. The shaded region shows standard error. \DirGPR{} consistently performs better than \BetaGP{}. \PerArmBeta{} is worse than its smoother counterparts.} 
\label{fig:estimation}
\end{figure*}

\begin{figure*}[t]
\centering
\begin{tabular}{cccc} \tabcolsep 1pt
\includegraphics[width=.22\hsize]%
{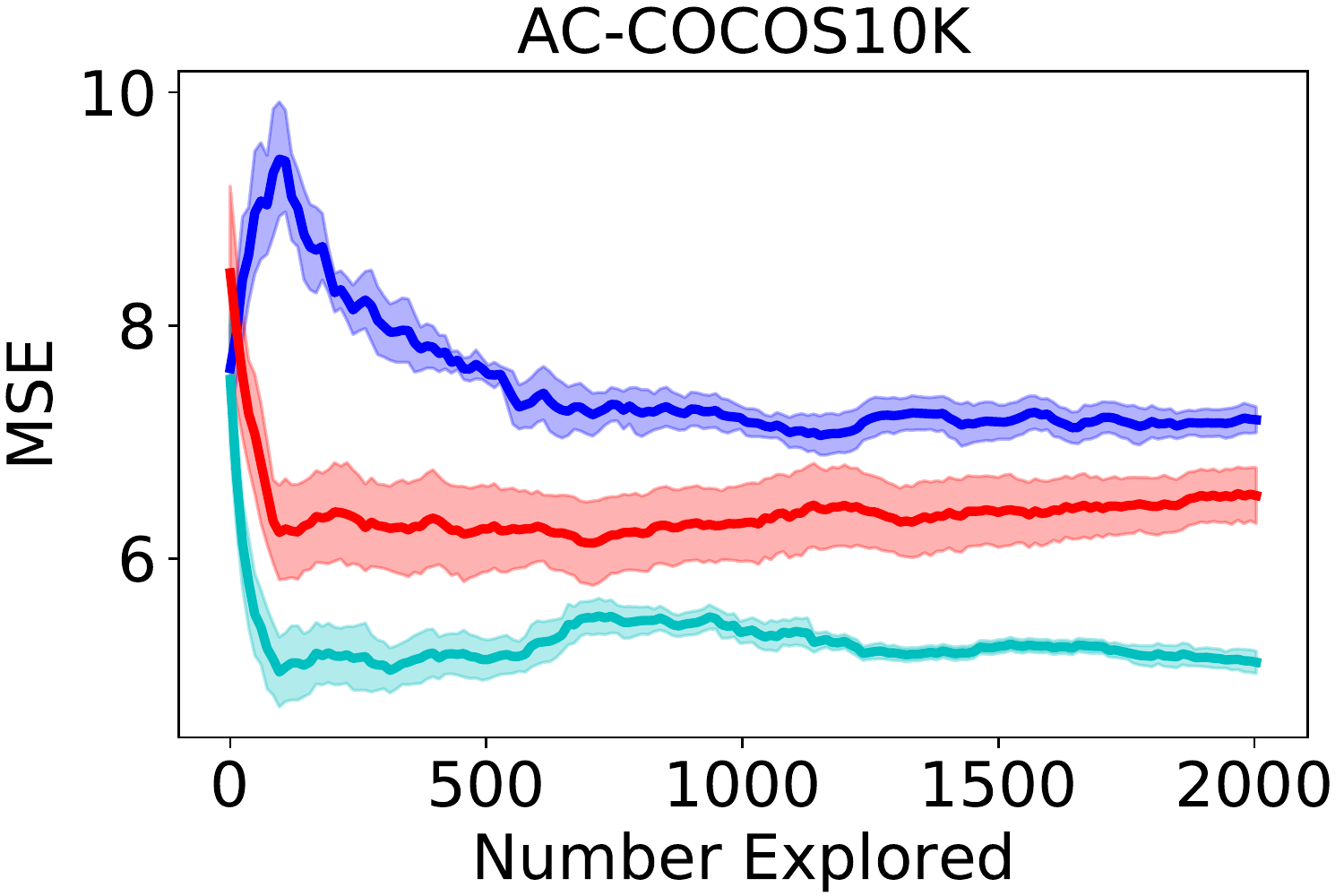} &
\includegraphics[width=.22\hsize]%
{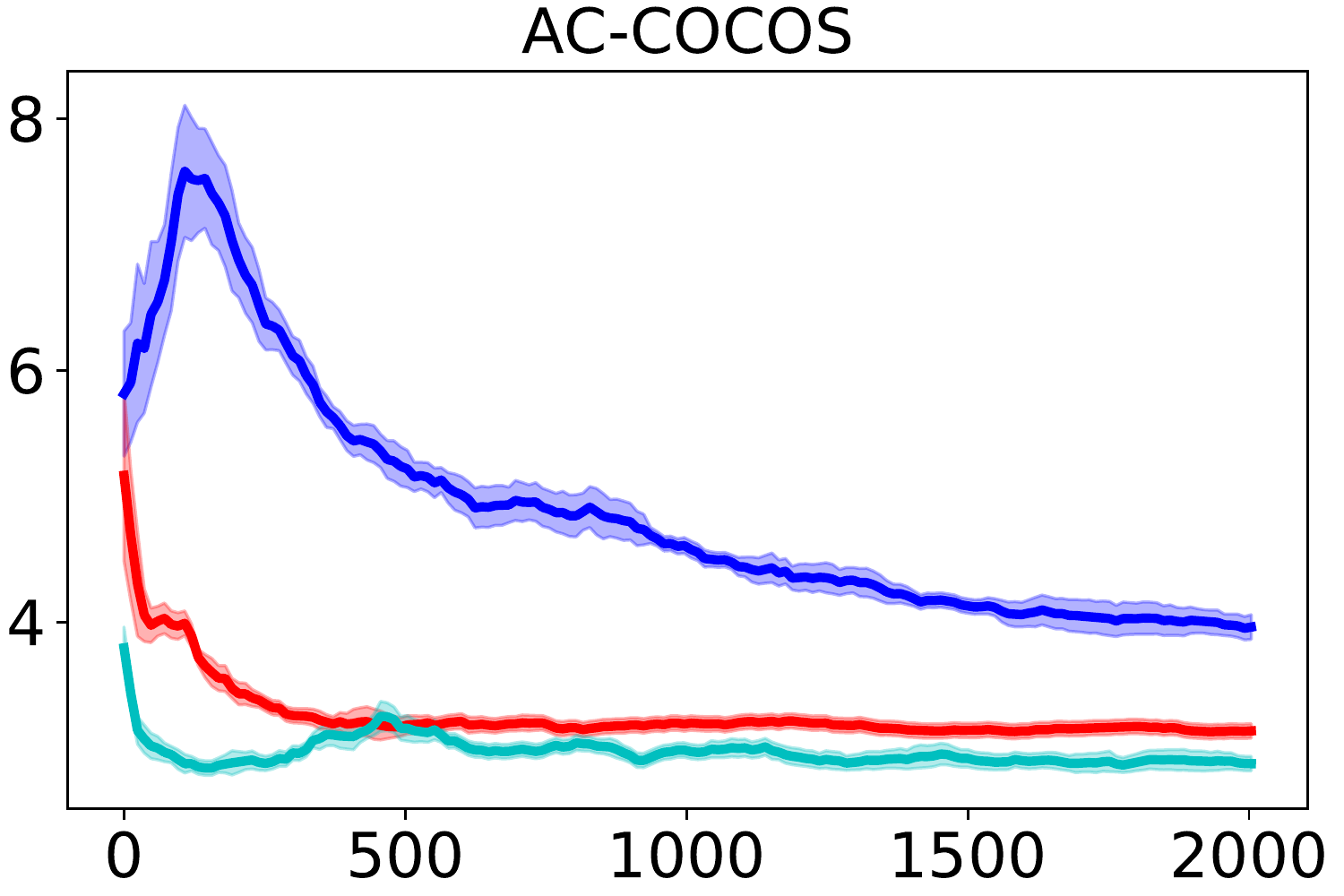} &
\includegraphics[width=.22\hsize]%
{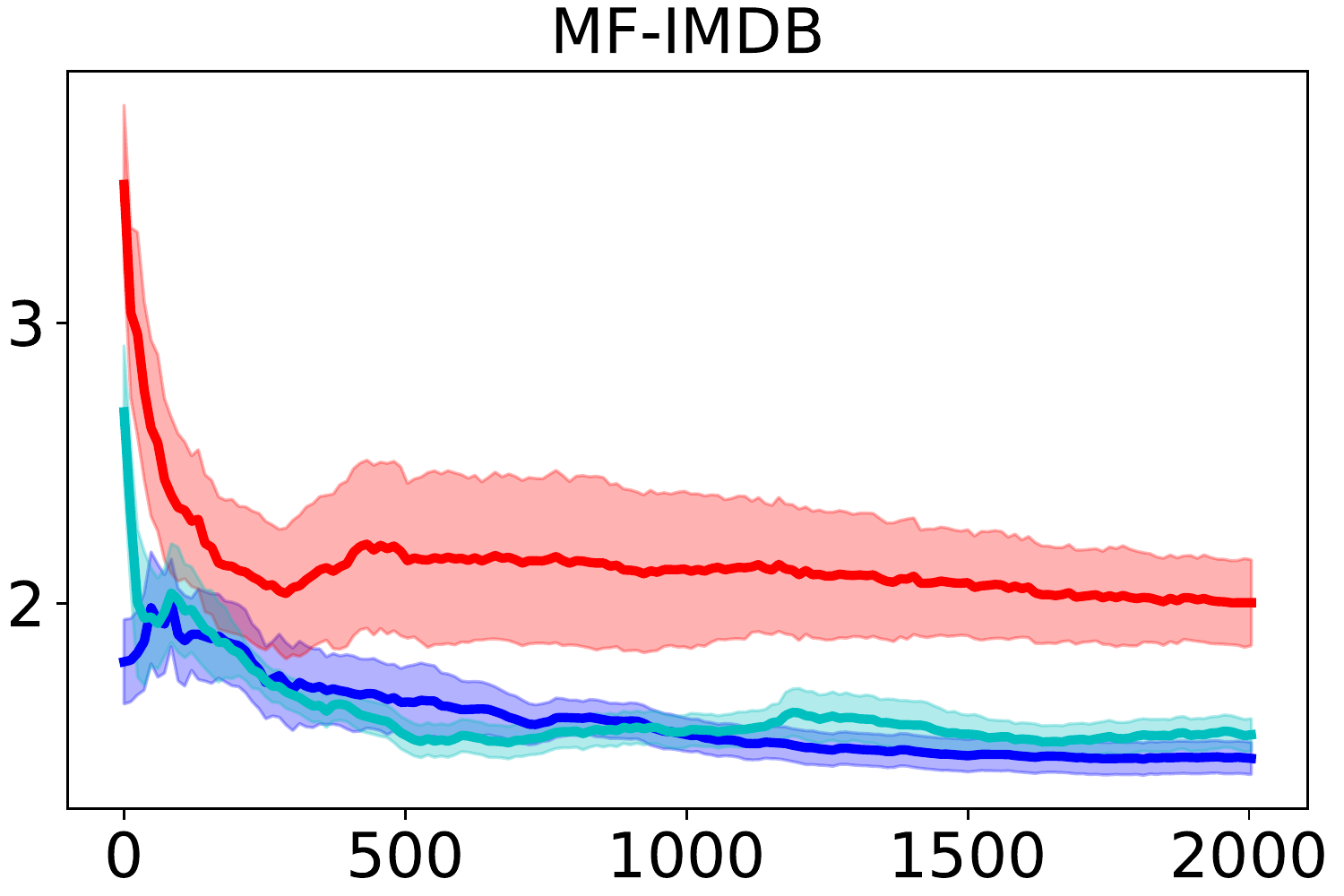} &
\includegraphics[width=.22\hsize]%
{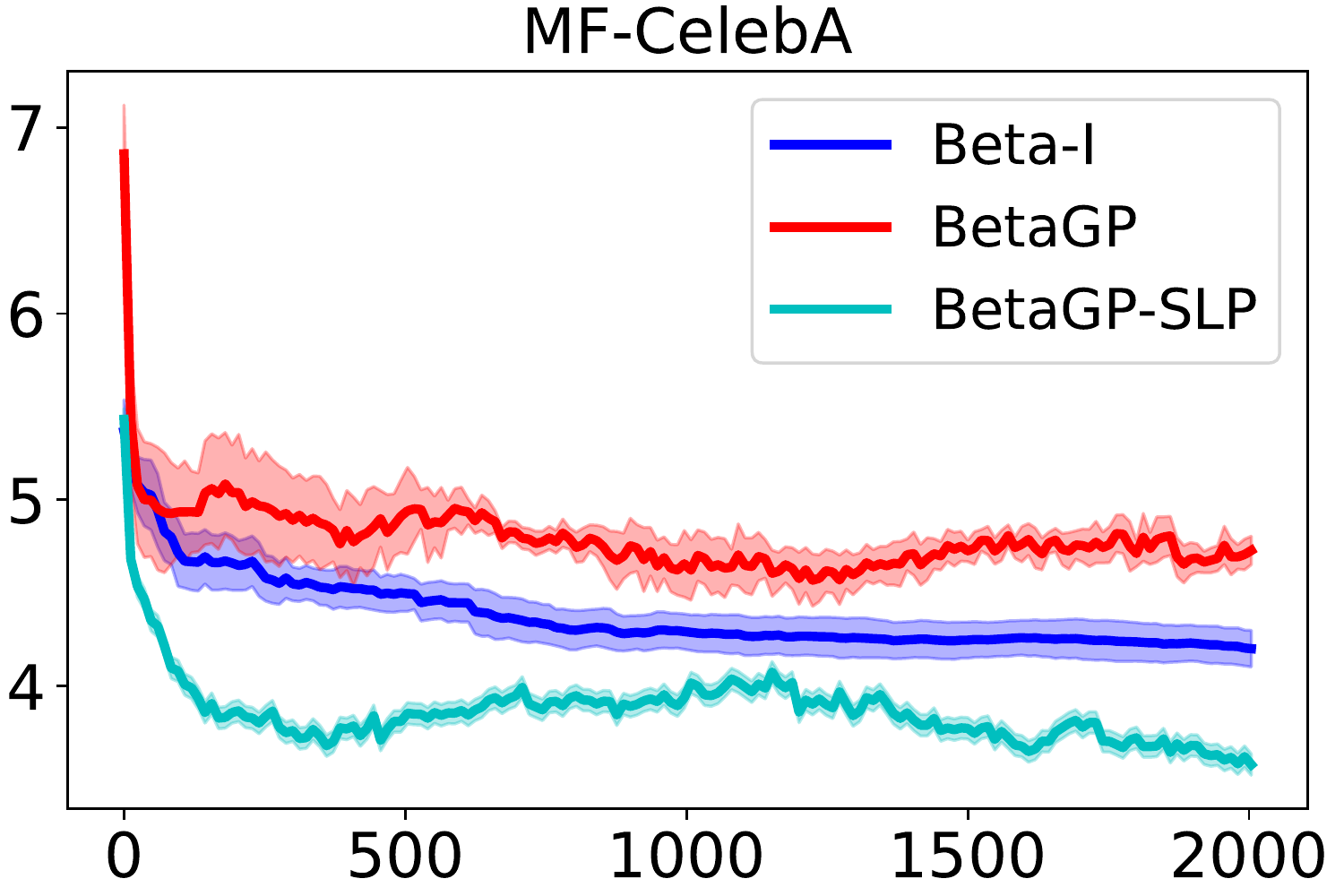}
\end{tabular}
\caption{Comparison of exploration methods. \DirGPR{} reduces macro MSE fastest most of the time. Shaded region shows standard error.} 
\label{fig:exploration}
\end{figure*}


\begin{figure*}
\centering
\begin{subfigure}{.24\textwidth}
  \centering
  \includegraphics[width=\linewidth]{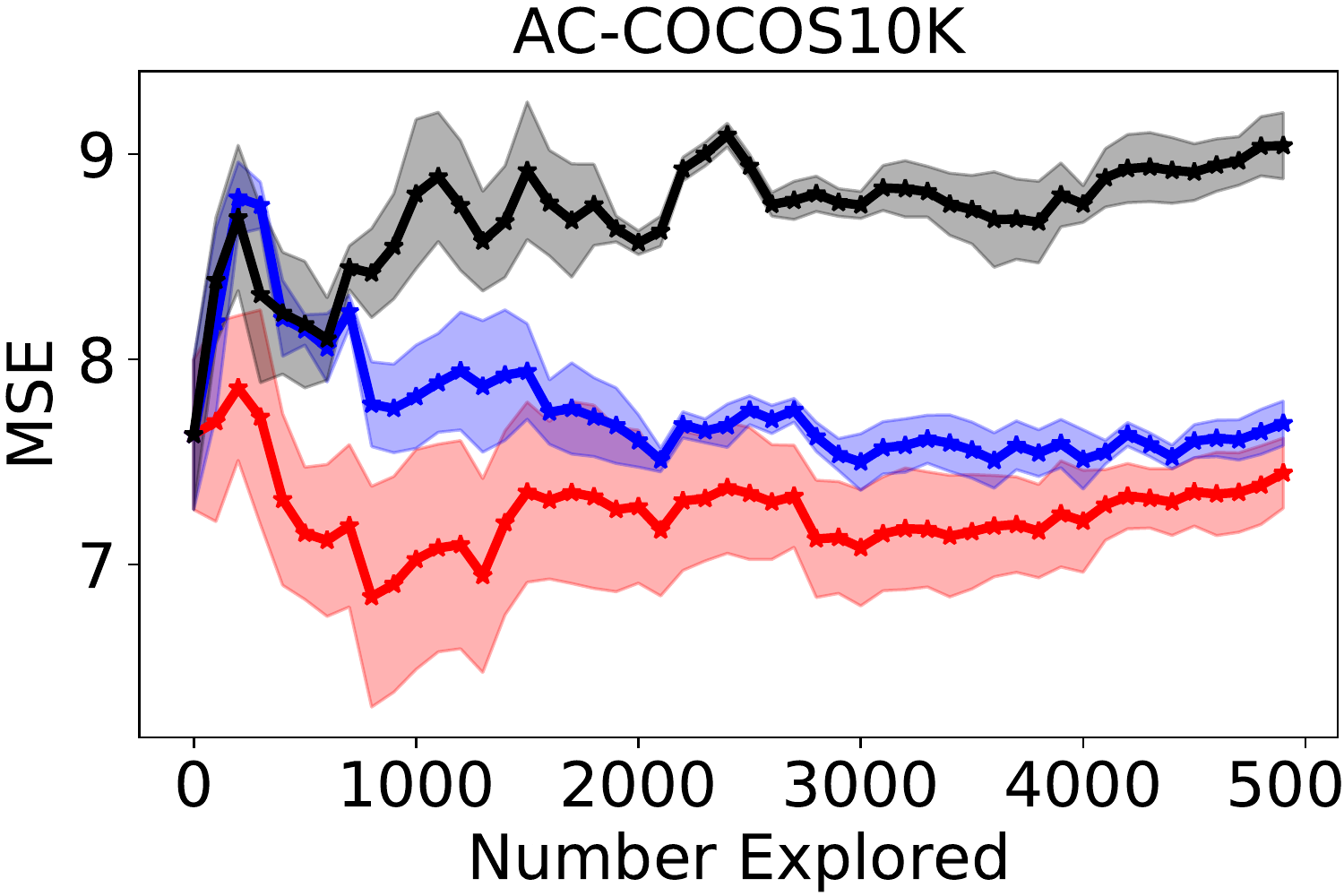}
\end{subfigure}
\begin{subfigure}{.25\textwidth}
  \centering
  \includegraphics[width=0.92\linewidth]{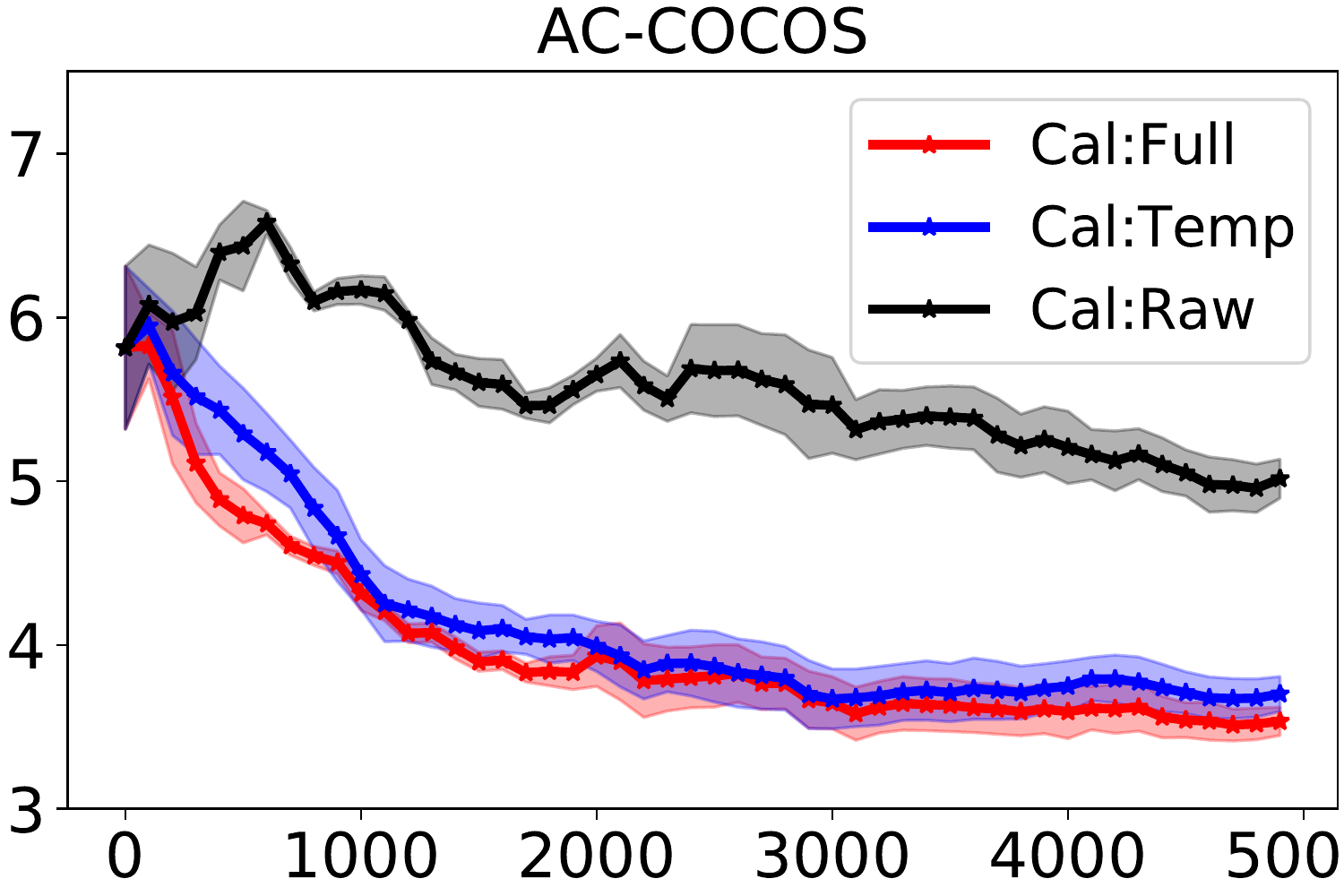}
\end{subfigure}
\begin{subfigure}{.24\textwidth}
  \centering
  \includegraphics[width=\linewidth]{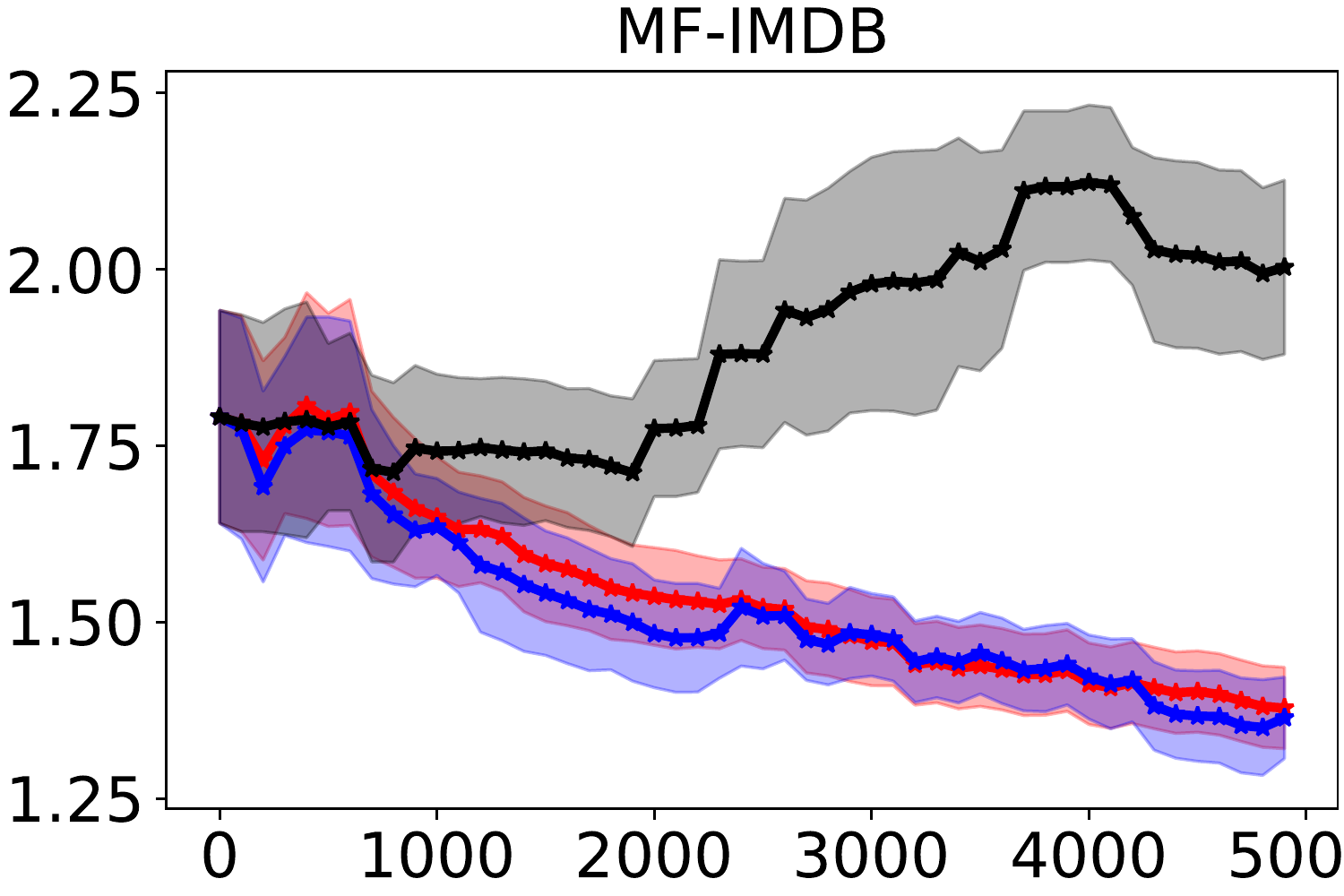}
\end{subfigure}
\begin{subfigure}{.24\textwidth}
  \centering
  \includegraphics[width=\linewidth]{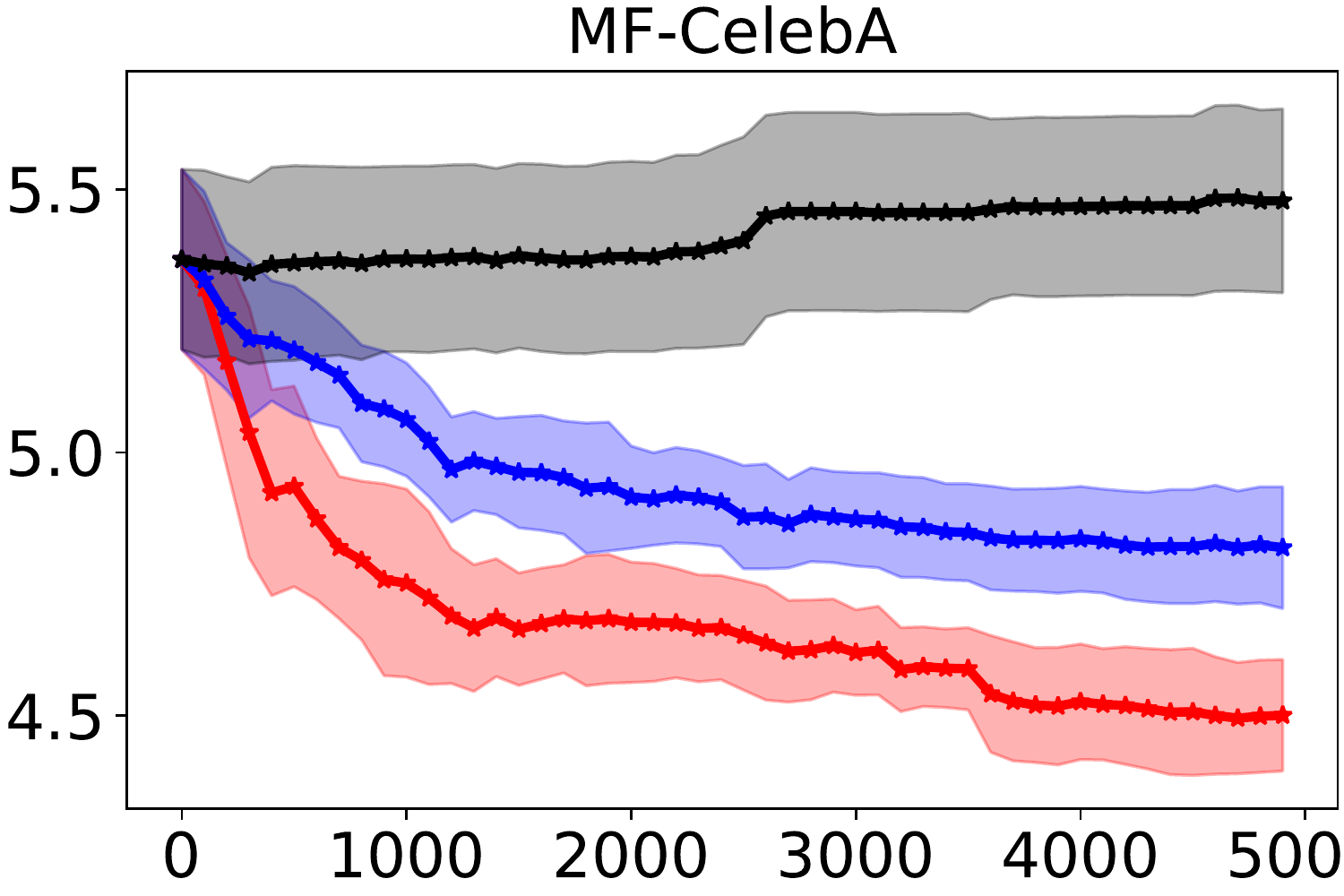}
\end{subfigure}%
\caption{Calibration methods compared on different tasks.  \calFull{} (red) includes temperature-based recalibration and correlation modeling with joint potential and gives the best macro MSE. Shaded region shows standard error.}
\label{fig:calibration}
\end{figure*}

\subsection{Accuracy Estimation Quality}
\label{sec:expt:est}
\vspace{-0.2cm}
We evaluate methods on their estimation quality when each method is provided with exactly the same (randomly chosen) labeled set. We compare the four service models when fitted on labeled data of size 2,000 and the results appear in Table~\ref{tab:est}. Note that we only have label supervision on $\cY$ in the labeled data. Table~\ref{tab:est} shows macro and worst MSE, standard deviation for each metric can be found in Appendix~\ref{sec:appendix:metrics}. In  Figure~\ref{fig:estimation}, we show worst MSE for a range of labeled data sizes along with their error bars. We make the following observations.
{\bf Smoothing helps:} Since we have a large number of arms, we expect \PerArmBeta{} to fare worse than its smooth counterparts (\BernGP{} and \BetaGP{}), especially on the worst arms. This is confirmed in the table.  In three out of four cases, \PerArmBeta{} method is worse than even the constant predictor \cpred\ on both metrics.
{\bf Modeling arm specific noise helps:} \BetaGP{} is better than \BernGP{} on almost all the cases in the table.  
{\bf Significant gains when the scale supervision problem of \BetaGP{} is fixed:} \DirGP{} is significantly better than \BetaGP{} in the table and figure. 
{\bf Our pooling strategy helps:} \DirGPR{} improves \DirGP{} over worst MSE without hurting  macro MSE as seen in the table and figure.

\subsection{Exploration Efficiency}
\label{sec:expt:exp}
\vspace{-0.2cm}
We compare different methods that use their own estimated variance for exploring instances to label (Section~\ref{sec:aaa:Explore}), as a function of the number of explored examples --- see Figure~\ref{fig:exploration}. In most cases, \DirGPR{} gives the smallest macro MSE, beating \PerArmBeta{} and \BetaGP{}. Note \PerArmBeta\ is the exploration method recently suggested in \cite{JiLR20}. We observe that \BetaGP\ provides very poor exploration quality, indicating that the uncertainty of arms is not captured well by just using two GPs. In fact, in many cases \BetaGP\ is worse than \PerArmBeta, even though we saw the opposite trend in estimation quality (Figure~\ref{fig:estimation}). These experiments brings out the significant role of  Dirichlet scale supervision and pooled observations in enhancing the uncertainty estimates at each arm.

\subsection{Impact of Calibration}
\label{sec:expt:calib}

\vspace{-0.2cm}
We consider two baselines along with our method explained in Section~\ref{sec:aaa:AttribNoise}: \textbf{\calRaw}, which uses the predicted attribute from the attribute models without any calibration and \textbf{\calTemp}, which  calibrates only the temperature parameters shown in eqn.~\eqref{eqn:factor:assume}, i.e., without the joint potential part. We refer to our method of calibration using temperature and joint potentials as \textbf{\calFull}.
%
%
We compare these on the four tasks with estimation method set to {\PerArmBeta} and random exploration strategy. Figure~\ref{fig:calibration} compares the three methods: \calRaw (Black), \calTemp (Blue), \calFull (Red).  
The X-axis is the number of explored examples beyond $\dseed$, and Y-axis is estimation error. Observe how \calTemp{} and \calFull\ are consistently better than \calRaw, and \calFull\ is better than~\calTemp.

\section{Related Work}
\label{sec:rel}

Our problem of actively estimating the accuracy {\em surface} of a classifier generalizes the more established problem of estimating a single accuracy {\em score} \citep{sawade10,Sawade2012,katariya12,druck11,bennett10,Karimi2020}.  For that problem, a known solution is stratified sampling, which partitions data into homogeneous strata and then seeks examples from regions with highest uncertainty and support. If we view each arm as a stratum, our method follows similar strategy. A key difference in our setting is that low support arms cannot be ignored. This makes it imperative to calibrate well the uncertainty under limited and skewed support distribution.  The setting of \citet{JiLR20} is the closest to ours. However, their work only considers a single attribute which they fit using \PerArmBeta, whereas we focus on the challenges of estimating accuracy over many sparsely populated attribute combinations.

{\bf Sub-population performance:} Several recent papers have focused on identifying sub-populations with significantly worse accuracy than aggregated accuracy
~\citep{Sagawa19,Oakden20hidden,WILDS20,JiSS20neurips,Miller21,Subbaswamy21eval}. Some of these have also proposed sample-efficient techniques~\citep{JiSS20neurips, Miller21} for estimation of performance on specific sub-groups, such as the ones defined by attributes like gender and race. Our accuracy surface estimation problem can be seen as a generalization where we need to estimate for all sub-groups defined in the Cartesian space of pre-specified semantic attributes. 
\citet{modelcards19} recommend reporting model performance under the influence of various relevant demographic/environmental factors as model cards--similar to the accuracy surface. 

{\bf Experiment design:} Another related area is experiment design using active explorations with GPs \citep{SrinivasKK09}.  Their goal is to find the mode of the surface whereas our goal is to estimate the entire surface.  Further, each arm in our setting corresponds to multiple instances, which gives rise to a degree of heteroscedasticity and input-dependent noise that is not modeled in their settings. \citet{LzaroGredilla2011VariationalHG,Kersting2007MostLH} propose to handle heteroscedasticity by using a separate GP to model the variance at each arm. However, we showed the importance of additional terms in our likelihood and observation pooling to reliably represent estimation uncertainty.
\citet{WengerKT20} propose observation pooling for estimating smooth Betas but they assume a fixed kernel.


{\bf Model debugging:} Testing deep neural network (DNN) is another related emerging area~\citep{Zhang2020}. 
\citet{PeiCY17DeepXplore,TianPJ18DeepTest,SunWR18Concolic, odena2019tensorfuzz} propose to generate test examples with good coverage over all activations of a DNN.
%
\citet{Ribeiro18Anchors,KimGP20} identify rules that explain the model predictions. 



\section {Conclusion}
\label{sec:end}

We presented \shortname, a new approach to estimate the accuracy of a classification service, not as a single number, but as a surface over a space of attributes (arms).  \shortname{} models uncertainty with a Beta distribution at each arm and regresses these parameters using two Gaussian Processes to capture smoothness and generalize to unseen arms.  We proposed an additional Dirichlet likelihood to mitigate an over-smoothing problem with GP's estimation of Beta distributions' scale parameters.
Further, to protect these high-capacity GPs from unreliable accuracy observations at sparsely populated arms, we propose to use  an observation pooling strategy. Finally, we show how to handle noisy attribute labels by an efficient joint recalibration method.
Evaluation on real-life datasets and classification services
show the efficacy of \shortname, both in estimation and exploration quality.\\
{\bf Limitation and future work:} (1)~We have evaluated \shortname{} on the order of thousands of arms.  Even larger attribute spaces could unearth more challenges. (2)~Identifying relevant attributes for an application can be non-trivial. Future work could devise strategies for attribute selection. 
(3)~Characterizing test-time data shifts could in itself be hard, particularly for text --- there could be subtle changes in word usage, style, or punctuation.  A more expressive attribute space needs to be developed for text applications.

\section{Acknowledgements}
The first author is supported by Google PhD Fellowship. This research was partly sponsored by the IBM AI Horizon Networks - IIT Bombay initiative.  

\bibliography{main}
\bibliographystyle{unsrtnat}

\clearpage

\appendix

{\centering \Large \bfseries \ztitle \\ (Appendix)
\par\bigskip}

\begin{table}[!h]
    \centering
    \begin{tabular}{|l|r|}
    \hline
        \thead{Main Section} & \thead{Appendix} \\\hline
        Source Code & Appendix~\ref{sec:appendix:code}\\\hline
        Section~\ref{sec:betagp} & Appendix~\ref{sec:appendix:betaab},~\ref{sec:appendix:gp} \\\hline
        Section~\ref{sec:aaa:sl} & Appendix~\ref{sec:appendix:synth}\\\hline
        Section~\ref{sec:aaa:AttribNoise} & Appendix~\ref{sec:appendix:calibration},~\ref{sec:appendix:pseudocode} \\\hline 
        Section~\ref{sec:expt:dataset},~\ref{sec:expt:services} & Appendix~\ref{sec:appendix:task},~\ref{sec:appendix:surface_stats}\\\hline
        Section~\ref{sec:expt:other},~\ref{sec:expt:est} & Appendix~\ref{sec:appendix:metrics} \\\hline
    \end{tabular}
    \caption{Mapping between main and appendix sections.}
    \label{tab:main_appendix_map}
\end{table}

\section{Source Code}
\label{sec:appendix:code}
Our code, dataset and instructions for replicating the results can be found at this \href{https://github.com/vps-anonconfs/aaa-neurips}{link}. 

\section{Parametric Form of \BetaGP{}}
\label{sec:appendix:betaab}

In Section~\ref{sec:betagp}, we claimed that \BetaGP{} with (mean, scale) parameterization is better than \BetaAB{} with the standard $(\alpha, \beta)$ parameterization of the Beta distribution. In this section, we present some empirical evidence corroborating the claim.

We compare between the two parametric forms with two service models in Table~\ref{tab:betaab}. For \BetaAB, we use two GPs, one to approximate the latent value corresponding to $\alpha$, and other for $\beta$. We use soft-plus operation  to transform the latent values to their admissible positive $\alpha$, $\beta$ values.

We report macro-averaged mean square errors on two tasks in Table~\ref{tab:betaab}, when fitting on 2,000 instances --- similar to the setting of Section~\ref{sec:expt:est}. We found the {\BetaAB} estimates unstable and far worse, perhaps because smoothness is not expected in either of $\alpha, \beta$ parameters across arms making the GP's bias ineffective.  

\begin{table}[H]
\centering
\begin{adjustbox}{max width=\textwidth}
\tabcolsep 2pt
\begin{tabular}{|l?{1.5pt}c|c?{1.5pt}c|c?{1.5pt}} \hline
Service$\rightarrow$ & \multicolumn{2}{c?{1.5pt}}{\bf MF-CelebA}  &  \multicolumn{2}{c?{1.5pt}}{\bf AC-COCOS} \\ \hline
Method$\rightarrow$ & \BetaGP\ & \BetaAB\ & \BetaGP\ & \BetaAB\ \\
1000 & 5.4 / 0.5 & 6.6 / 0.1 & 3.7 / 0.2 & 5.6 / 0.6\\
2000 &  4.6 / 0.8 & 6.2 / 0.1 & 3.3 / 0.2 & 4.8 / 0.1\\
3500 &  4.6 / 0.3 & 6.1 / 0.1 & 3.2 / 0.2 & 5.2 / 0.4\\\hline
\end{tabular}
\end{adjustbox}
\caption{Comparison of estimation error between \BetaGP{} with (mean, scale) parameterization vs.\ \BetaAB.  \BetaAB{} is worse than {\BetaGP}. }
\label{tab:betaab}
\end{table}

\section{More Details of Gaussian Process (GP) Setup}
\label{sec:appendix:gp}
In this section, we give further details on GP training,  posterior approximation and computational cost. This section elaborates on Section~\ref{sec:betagp}. 

In all our proposed estimators, the data likelihood is modeled either by a Bernoulli or a Beta distribution. Data likelihood term of ~\BernGP{}, ~\BetaGP{}, are shown in Eqn.~\eqref{eq:bernLL}, Eqn.~\eqref{eq:pointLLSimple}, respectively. Due to the non-Gaussian nature of the data likelihood, the posterior on parameters cannot be expressed in a closed form. Several approximations exist for fitting the posterior especially for the more standard \BernGP{}, we will discuss one such method in what follows. Recall that we model two latent values $f, g$ each modeled by an independent GP. They can be seen to have been drawn from a single GP with even larger dimension and with appropriately defined kernel matrix. For the sake of explanation and with a slight abuse of notation, we denote by $\vf$, the concatenation of $f$ and~$g$.  The corresponding kernel for the concatenated vector is appropriately made by combining the kernels of either of the latent values with kernel entries corresponding to interaction between $f$ and $g$ set to~0.

Variational methods are popular for dealing with non-Gaussian likelihoods in GP. In this method, we fit a multi-variate Gaussian that closely approximates the posterior, i.e. minimizes $\mathcal{D}_\text{KL}(q(\vf)\|P(\vf|D)$. GPs are often used in their sparse avatars using \emph{inducing points}~\citep{WilsonHS16,NickischR08} that provide approximations to the full covariance matrix with large computation benefits. As a result, q(f) is parameterized by the following trainable parameters (let `d', `m' denote the input dimension and number of inducing points resp.): (a) Z, a matrix of size $d\times m$, of locations of 'm' inducing points (b) $\mu\in \mathbb{R}^m, \Sigma\in \mathbb{R}^{m\times m}$, denoting mean and covariance of the inducing points. 
%
In order to minimize $\mathcal{D}_\text{KL}(q(\vf)\|P(\vf|D)$, a pseudo objective called Evidence Lower Bound (ELBO), shown below, is employed:
\begin{equation}
  q^*(\vf) = \argmax_{\vf \sim q(\vf)} \mathbb{E}_q[\log P(D|\vf)]-\mathcal{D}_\text{KL}(q(\vf)|| P(\vf))
  \label{eqn:elbo}
\end{equation}

The first term above in Eqn.~\eqref{eqn:elbo} maximizes data likelihood, which is Equation~\ref{eq:pointLLSimple} in our case.
The second term is a regularizer that regresses the posterior fit $q(\vf)$ close to the prior distribution $P(\vf)$ which is set to standard Normal. 
We optimize using this objective over all the parameters involved through gradient descent. The required integrals in \eqref{eqn:elbo} can be computed using Monte Carlo methods~\citep{HensmanMG15}. We describe further implementation details in the next section.

{\bf Implementation Details} \\
We use routines from GPytorch\footnote{\url{https://gpytorch.ai/}} library to implement the variational objective. Specifically, we extend \href{https://docs.gpytorch.ai/en/stable/models.html#gpytorch.models.ApproximateGP}{ApproximateGP} with \href{https://docs.gpytorch.ai/en/stable/variational.html?highlight=variationalstrategy#id1}{VariationalStrategy}, both of which are GPytorch classes, and set them to learn inducing point locations.

Number of \emph{inducing points} when set to a very low value could overly smooth the surface and can have high computation overhead when set to a large value. We set the number of inducing points to 50 for all the tasks. The choice of 50 over a larger number is only to ensure reasonable computation speed. 

Since we keep getting more observations as we explore, we use the following strategy for scheduling the parameter updates. We start with the examples in the seed set $\dseed$ and update for 1,000 steps. We explore using the variance of the estimated posterior. We pick 12 arms with highest variance and label one example for each arm. After every new batch of observations, we make 50 update steps on all the data. As a result, we keep on updating the parameters as we explore more. We use Adam Optimizer with learning rate $10^{-3}$. At each step, we update over observations from all the arms. The flow of the exploration and parameter update is also shown in Algorithm~\ref{alg:main}.

We use the feature representations of the network used to model joint potentials described in Section~\ref{sec:aaa:AttribNoise} to also initialize the deep kernel induced by~$\Emb$. A final new linear layer of default output size 20 is added to project the feature representations. 

In our proposed method~\DirGPR{}, described in Section~\ref{sec:aaa:pool}, we take the kernel average of three neighbours for any arm with fewer than five observations. 

\section{Simple Setting}
\label{sec:appendix:synth}

In Section~\ref{sec:aaa:sl}, we describe how the objective of \BetaGP{} does not supervise the scale parameter. Further, in Section~\ref{sec:aaa:pool}, we posit that the presence of sparse observations leads to learning a non-smooth kernel. In this section, we illustrate these two observations using a simple setting. 

We consider a simple estimation problem with 10 arms, their true accuracies go from 0.1 to a large value of 0.9 and then back to a small value of 0.2 as shown in the Table~\ref{tab:simple:fit}. In Table~\ref{tab:simple:fit}, we also show the number of observations per arm; Observe that the first three and the last three arms are sparsely observed.


\begin{table}[th]
    \centering
    \setlength{\tabcolsep}{4pt}
    \begin{tabular}{|l|r|r|r|r|r|r|r|r|r|r|}
        \hline
         \thead{Arm Index} &  \thead{1} & \thead{2} & \thead{3} & \thead{4} & \thead{5} & \thead{6} & \thead{7} & \thead{8} & \thead{9} & \thead{10}\\\hline
         Accuracy & 0.1 & 0.3 & 0.5 & 0.7 & 0.8 & 0.9 & 0.6 & 0.4 & 0.3 & 0.2 \\\hline
         N & 1 & 1 & 1 & 20 & 20 & 20 & 20 & 1 & 1 & 1 \\\hline
         \multicolumn{11}{c}{Estimated Scale Value}\\\hline
         \BetaGP{}  & 10.33 & 10.88 & 11.37 & 11.73 & 11.92 & 11.91 & 11.72 & 11.34 & 10.85 & 10.29 \\
         \DirGP{} &  1.49 & 1.42 & 1.59 & 9.64 & 10.40 & 10.42 & 9.58 & 1.59 & 1.42 & 1.49\\ 
         \DirGPR{} &  1.72 & 1.63 & 1.92 & 9.63 & 10.38 & 10.27 & 9.48 & 1.93 & 1.53 & 1.62\\\hline
    \end{tabular}
    \caption{Arms, their indices, accuracies and number of observations (N) in the simple setting are shown in first three columns in that order. The scale parameter estimated using one of the algorithms for each arm is shown in the last three columns. Notice that \BetaGP{} fitted scale parameter does not reflect the underlying observation sparsity for the first and last three arms.}
    \label{tab:simple:fit}
\end{table}

We now present the fitted values by some of the methods we discussed in the main section. The index of an arm is the input for any estimator with no feature learning. Our motivation for discussing the simple setting is to illustrate the two limitations we discussed in the main content regarding the \BetaGP{} objective: (a) the scale parameter of the \BetaGP{} objective is not supervised (b) sparse observations lead to non-smooth surface. Toward these ends, we evaluate \BetaGP{},~\DirGP{},~\DirGPR{} methods on this setting. The fitted scale parameters for each arm by each of the estimators is shown in 
Table~\ref{tab:simple:fit}. Observe that \BetaGP{} fitted scale parameter does not reflect the underlying observation sparsity of the first, last three arms. 
\DirGP{},~\DirGPR{} fitted scale values more faithfully reflect the underlying number of observations. All the numbers reported here are averaged over 20 seed runs. 

\begin{table}[ht]
    \centering
    \begin{tabular}{|l|r|r|r|}
    \hline
         Method & \thead{Bias$^2$} & \thead{Variance} & MSE \\\hline
         \BetaGP{} & 0.052 & 1.011 & 1.063 \\
         \DirGP{} & 0.051 & 1.010 & 1.061 \\
         \DirGPR{} & 0.095 & 0.221 & {\bf 0.316} \\\hline
    \end{tabular}
    \caption{Bias-variance decomposition of MSE in the simple setting}
    \label{tab:simple:bv}
\end{table}

In Table~\ref{tab:simple:bv}, we show the bias$^2$, variance decomposition of the mean squared error from 20 independent runs. Observe that ~\BetaGP{},~\DirGP{} have low bias but large variance and \DirGPR{} has much lower variance at a slight expense of bias, as a result the overall MSE value for \DirGPR{} is much lower than the other two. Moreover, we look at the fitted kernel length parameter (recall from Equation~\eqref{eq:kernel}) as a proxy for smoothness of the fitted kernel. Large kernel length is indicative of long range smoothness. The average kernel length for \BetaGP{},~\DirGP{},~\DirGPR{} are 0.67, 0.68, 1.87 respectively. Despite using a GP kernel, we find the estimates of \BetaGP{},~\DirGP{} of high variance, that is also indicative of short range smoothness apparent from the low average kernel length. On the other hand, \DirGPR{} imposes long range smoothness, as a result, decreases the MSE value more effectively when compared with \DirGP{}.

\section{Calibration Training Details}
\label{sec:appendix:calibration}

In this section, we give further training details on the noise calibration method discussed in Section~\ref{sec:aaa:AttribNoise}.

As discussed in Section~\ref{sec:aaa:AttribNoise}, we use both labeled, small $\dseed$ and large $U$ for training calibration parameters that are expressed in the objective~\eqref{eqn:max_likelihood}.
On the unlabeled data $U$, we use attribute values predicted using the predictors: $\{\attpred_k \mid k \in \attlist\}$ as a proxy for true values. The use of predicted values as the replacement for true value under-represents the attribute prediction error rate and interferes in the estimation of temperature parameters $t$. However, if we use $U$ for training, we see a lot more attribute combinations and this can help identify more natural attribute combinations aiding in the learning of joint potential parameters of $N$. 
We mitigate the temperature estimation problem by up-sampling instances in $\dseed$ such that the loss in every batch contains equal contribution from $\dseed$ and~$U$. 
 
Recall that the MLE objective~\eqref{eqn:max_likelihood}, contains contribution from two terms: (a)~temperature scaled logits (b)~attribute combination potential. In practice, we found that the second term (b) dominates the first, this causes under-training of the temperature parameters. Ideally, the two terms should be comparable and replaceable. We address this issue by dropping the second term corresponding to the network-assigned edge potential term in the objective half the times, which estimates better the temperature parameters. Further, we use a small held out fraction of $\dseed$ for network architecture search on~$L$, and for early stopping. The training procedure is summarized in Alg.~\ref{alg:calibrate}.

\section{More Task and Dataset Details}
\label{sec:appendix:task}

\subsection{MF-CelebA, MF-IMDB}
As mentioned in Section~\ref{sec:expt:dataset}, we hand-picked 12 binary attributes relevant for gender classification of the 40 total available attributes in the CelebA dataset. 
The twelve binary attributes are listed in Table~\ref{tab:celeba:attrs}, these constitute the $\attlist$. $\attspace$ is the combination of twelve binary attributes and is $2^{12}=4,096$ large. 
The attributes related to hair color are retained in this list due to the recent finding that hair-color is spuriously correlated with the gender in CelebA~\citep{Sagawa19}. We ignored several other gender-neutral or rare attributes. 

\begin{table}[htb]
    \centering
    \begin{tabular}{|l|r|r|}\hline
    \thead{Index} & \thead{Name} & \thead{Num. labels} \\\hline
    1 & Black Hair? & 2\\\hline
    2 & Blond Hair? & 2\\\hline
    3 & Brown Hair? & 2\\\hline
    4 & Smiling? & 2\\\hline
    5 & Male? & 2\\\hline
    6 & Chubby? & 2\\\hline
    7 & Mustache? & 2\\\hline
    8 & No Beard? & 2\\\hline
    9 & Wearing Hat? & 2\\\hline
    10 & Blurry? & 2\\\hline
    11 & Young? & 2\\\hline
    12 & Eyeglasses? & 2\\\hline
    \end{tabular}
    \caption{Attribute list of MF-CelebA, MF-IMDB.}
    \label{tab:celeba:attrs}
\end{table}

\subsection{AC-COCOS, AC-COCOS10K}

COCOS is a scene classification dataset, where pixel level supervision is provided. Methods are usually evaluated on pixel level classification accuracy. For simplicity, we cast it in to an object recognition task. The subset of ten animal labels we consider is shown in Table~\ref{tab:cocos:primary}. We consider in our task five coarse background (stuff) labels by collapsing the fine labels to coarse using the label hierarchy shown in Table~\ref{tab:cocos:stuff} and is the same as the official hierarchy~\citep{COCOS}.

We now describe how we cast the scene label classification task to an animal classification task. We first identify the subset of images in the train and validation set of the COCOS dataset which contain only one animal label, it could contain multiple background labels. If the image contains multiple animals, we exclude it, leaving around 23,000 images in the dataset. This is implemented in the routine: \texttt{filter\_ids\_with\_single\_object} of \texttt{cocos3.py} in the attached code.  We also retrofit the scene classification models for animal classification. When the model (service model) labels pixels with more than one animal label, we retain the label associated with the largest number of pixels. This is implemented in \texttt{fetch\_preds} routine of \texttt{cocos3.py}. Recall our calibration method makes use of the logit scores given by the attribute predictors, since we are aggregating prediction from multiple pixels, we do not have access to the logit scores. We simply set the logit score to +1 if a label is found in the image and -1 otherwise. 

We follow the same protocol for both the tasks: AC-COCOS, AC-COCOS10K. The only difference between the two is the service model, AC-COCOS is a stronger service model trained on 164K size training data compared to AC-COCOS10K which is a model trained on a previous version of the dataset that is only 10K large. In these tasks, we use the same model for predicting the attributes and task labels since the pre-trained model we use is a scene label classifier. Both the pretrained\footnote{\url{https://github.com/kazuto1011/deeplab-pytorch}} models were trained using ResNet101 architecture.  

\begin{minipage}{0.2\textwidth}
    \centering
    \begin{tabular}{|l|}\hline
    \thead{Name}\\\hline
    bird\\\hline
    cat\\\hline
    dog\\\hline
    horse \\\hline
    sheep \\\hline 
    cow \\\hline
    elephant \\\hline 
    bear \\\hline 
    zebra \\\hline
    giraffe \\\hline
    \end{tabular}
    \captionof{table}{List of ten animals in AC tasks}
    \label{tab:cocos:primary}
\end{minipage}
\hfill
\begin{minipage}{0.8\textwidth}
    \centering
    \begin{tabular}{|l|l|}\hline
         \thead{Coarse label} & Example of constituent stuff classes \\\hline
         water-other & sea, river \\\hline
        ground-other & ground-other, playingfield, platform, railroad, pavement \\\hline
        sky-other & sky-other, clouds \\\hline
        structural-other & structural-other, cage, fence, railing, net \\\hline
        furniture-other & furniture-other, stairs, light, counter, mirror-stuff \\\hline
    \end{tabular}
    \captionof{table}{80 stuff labels in the COCOS dataset are collapsed in to five coarse labels. Few examples are shown for each coarse label in the right column.}
    \label{tab:cocos:stuff}
\end{minipage}


\section{Statistics of Accuracy Surface}
\label{sec:appendix:surface_stats}
We show in Table~\ref{tab:dataset:stats}, details about our two data sets such as the number of attributes, number of arms and number of active arms. Active arms are arms with a support of at least five and are the ones used for evaluation. Large number of arms as shown in the table exclude the possibility of manual supervision, since it is hard to obtain and label data that covers all the arms.  

\begin{figure}
    \centering
    \begin{subfigure}[t]{0.48\textwidth}
    \vspace{0pt}
        \includegraphics[width=\textwidth]{images/quantiles_new.pdf}
        \caption{We show the mean and ten quantiles of per-arm accuracy: 0, 0.1, 0.3, 0.5, 0.7, 0.9, 1. for each task when evaluated on their corresponding dataset (quantile 0 corresponds to the worst value). Observe the disparity between the best and the worst arms in terms of accuracy. In all the cases, also note how the large mean accuracy (macro-averaged over arms) does not do justice to explaining the service model's vulnerabilities.}
    \label{fig:task:stats1}
    \end{subfigure}
    \hfill
    \begin{subfigure}[t]{0.48\textwidth}\vspace{0pt}
        \includegraphics[width=\textwidth]{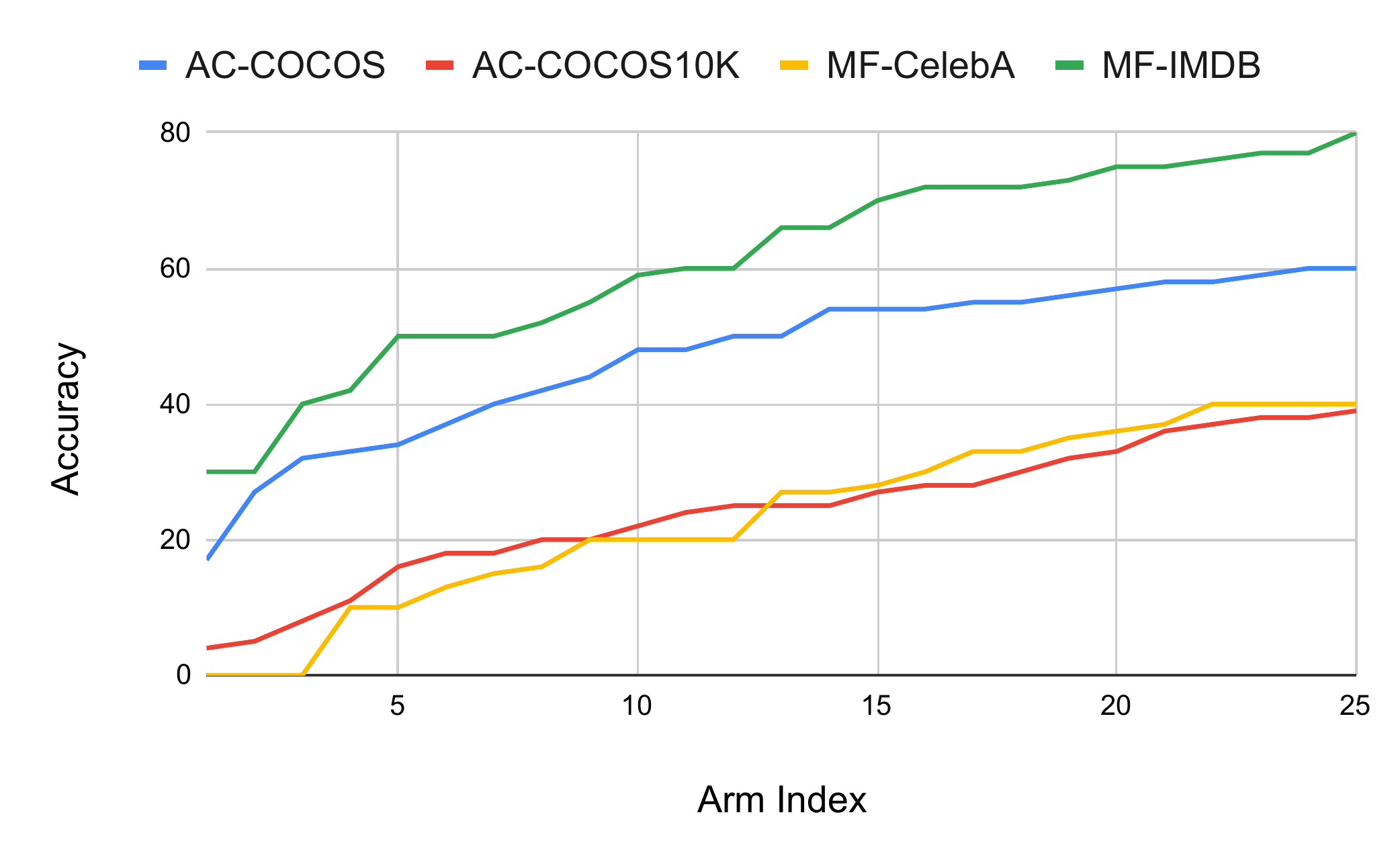}
        \caption{We show here the 25 worst arm accuracies for each of the service models. The large number of arms with accuracies much worse than mean accuracy further illustrates our argument for why we need accuracy surfaces.}
        \label{fig:task:stats2}
    \end{subfigure}
    \caption{Arm accuracies.}
    \label{fig:task}
\end{figure}

In Figures~\ref{fig:task:stats1} and~\ref{fig:task:stats2}, we show some statistics that illustrate the shape of the accuracy surface. We note that, although the service model's mean accuracy is high, the accuracy of the arms in the 10\% quantile is abysmally low while arms in the top-quantiles have near perfect accuracy. This further motivates for why we need an accuracy surface instead of single accuracy estimate.  

\begin{table}[H]
\begin{center}
\begin{tabular}{|l|r|r|r|}
\hline
Dataset & \# attributes & \# arms & \# active arms \\\Xhline{4\arrayrulewidth}
CelebA & 12 & 4096 & 398\\\hline
COCOS & 6 & 320 & 176 \\\hline
\end{tabular}
\end{center}
\caption{Attribute statistics per dataset. First and second column show number of attributes and total possible combinations of the attributes. Third column shows number of attribute combinations (arms) with at least a support of five in the unlabeled data. These are the arms on which accuracy surface is evaluated.}
\label{tab:dataset:stats}
\end{table}

\section{Standard Deviation and More Evaluation Metrics}
\label{sec:appendix:metrics}
We include standard deviation accompanying numbers in Table~\ref{tab:est} for macro MSE in Table~\ref{tab:macromse} and for worst MSE in Table~\ref{tab:worstmse}.

In the main content of the paper, we gave results using macro and worst MSE. In this following section, we show results using two other metrics.  We follow the same setup as in Section~\ref{sec:expt:est}.

{\bf Micro-averaged MSE:} We assign importance to each arm based on its support. The error per arm is multiplied by its support (in $U$). Results with this error are shown in Table~\ref{tab:micromse}. The best predictor with this metric is the point estimate given by the \cpred{} estimator which is not surprising since very few arms with high frequency dominate this metric.\\
{\bf Infrequent MSE:} In Table~\ref{tab:infreqmse}, we show MSE evaluated only on the 50 arms that are least frequent in $U$. \\
For each of the above evaluation metrics, the trend between \BetaGP\, \DirGP{}, \DirGPR\ is statistically significant. 

\begin{table}
\centering
\begin{tabular}{|l|llllll|}\hline
Service $\downarrow$ &\thead{\cpred} & \thead{\PerArmBeta} & \thead{\BernGP{}} & \thead{\BetaGP{}} & \thead{\DirGP{}} & \thead{\DirGPR{}} \\\hline
AC-COCOS10K & 5.4 / 0.2 & 7.0 / 0.6 & 7.0 / 0.7 & 7.1 / 0.3 & 5.3 / 0.2 & 4.7 / 0.1\\
AC-COCOS & 3.2 / 0.1 & 4.3 / 0.3 & 3.5 / 0.3 & 3.3 / 0.2 & 2.8 / 0.0 & 2.8 / 0.1\\
MF-IMDB & 1.2 / 0.0 & 1.6 / 0.1 & 1.7 / 0.2 & 2.2 / 0.2 & 1.4 / 0.1 & 1.4 / 0.1\\
MS-CelebA & 5.2 / 0.1 & 4.7 / 0.2 & 4.9 / 0.4 & 4.6 / 0.8 & 4.1 / 0.1 & 4.3 / 0.1\\\hline
\end{tabular}
\caption{\emph{Macro-averaged} MSE along with standard deviation on all tasks. Shown after trailing '/' is the standard deviation.}
\label{tab:macromse}
\end{table}

\begin{table}
\centering
\begin{tabular}{|l|llllll|}\hline
Service $\downarrow$ &\thead{\cpred} & \thead{\PerArmBeta} & \thead{\BernGP{}} & \thead{\BetaGP{}} & \thead{\DirGP{}} & \thead{\DirGPR{}} \\\hline
AC-COCOS10K & 3.0 / 0.0 & 3.4 / 0.2 & 3.6 / 0.2 & 3.5 / 0.1 & 3.2 / 0.1 & 3.3 / 0.4\\
AC-COCOS & 1.4 / 0.0 & 1.8 / 0.2 & 1.6 / 0.1 & 1.6 / 0.1 & 1.7 / 0.1 & 2.1 / 0.3\\
MF-IMDB & 0.2 / 0.0 & 0.3 / 0.1 & 0.2 / 0.0 & 0.3 / 0.0 & 0.3 / 0.1 & 0.8 / 0.1\\
MF-CelebA & 0.8 / 0.0 & 0.7 / 0.1 & 0.7 / 0.1 & 0.7 / 0.2 & 0.9 / 0.1 & 1.2 / 0.1\\\hline
\end{tabular}
\caption{\emph{Micro-averaged} MSE along with standard deviation on all tasks. Shown after trailing '/' is the standard deviation.}
\label{tab:micromse}
\end{table}

\begin{table}
\centering
\begin{tabular}{|l|llllll|}\hline
Service $\downarrow$ &\thead{\cpred} & \thead{\PerArmBeta} & \thead{\BernGP{}} & \thead{\BetaGP{}} & \thead{\DirGP{}} & \thead{\DirGPR{}} \\\hline
AC-COCOS10K & 15.0 / 0.8 & 15.6 / 0.3 & 13.2 / 2.2 & 14.3 / 3.0 & 11.7 / 1.7 & 10.4 / 1.5\\
AC-COCOS & 9.4 / 0.4 & 10.0 / 0.4 & 8.6 / 0.7 & 7.9 / 0.9 & 6.8 / 0.5 & 5.7 / 0.5\\
MF-CelebA & 8.2 / 0.2 & 8.4 / 0.7 & 7.6 / 1.3 & 6.6 / 0.7 & 4.4 / 0.6 & 3.9 / 0.7\\
MF-IMDB & 35.9 / 0.6 & 30.3 / 1.2 & 28.1 / 2.7 & 25.9 / 2.7 & 22.6 / 1.4 & 23.3 / 2.3\\\hline
\end{tabular}
\caption{\emph{Worst} MSE along with standard deviation on all tasks. Shown after trailing '/' is the standard deviation.}
\label{tab:worstmse}
\end{table}

\begin{table}
\centering
\begin{tabular}{|l|llllll|}\hline
Service $\downarrow$ &\thead{\cpred} & \thead{\PerArmBeta} & \thead{\BernGP{}} & \thead{\BetaGP{}} & \thead{\DirGP{}} & \thead{\DirGPR{}} \\\hline
AC-COCOS10K & 7.3 / 0.3 & 11.8 / 1.5 & 10.8 / 1.9 & 12.4 / 0.1 & 7.3 / 0.2 & 6.4 / 0.4\\
AC-COCOS & 4.0 / 0.1 & 6.9 / 1.2 & 4.8 / 0.3 & 4.9 / 0.3 & 3.8 / 0.1 & 3.8 / 0.3\\
MF-CelebA & 2.9 / 0.0 & 2.8 / 0.0 & 3.3 / 1.2 & 3.9 / 0.4 & 3.2 / 0.2 & 2.9 / 0.3\\
MF-IMDB & 11.7 / 0.2 & 11.3 / 0.3 & 11.4 / 0.6 & 11.0 / 0.7 & 9.3 / 1.1 & 9.4 / 0.6\\\hline
\end{tabular}
\caption{\emph{Infrequent} MSE along with standard deviation on all tasks. Shown after trailing '/' is the standard deviation.}
\label{tab:infreqmse}
\end{table}

\section{Pseudocode}
\label{sec:appendix:pseudocode}
The full flow of estimation and exploration is summarized in Algorithm~\ref{alg:main}. Algorithm~\ref{alg:calibrate} shows the calibration training sub-routine. 

\begin{algorithm}
\caption{AAA: Accuracy Surface Estimator and Active Sampler}
\begin{algorithmic}[1]
\Require $\dseed , U, \{\attpred_k, k\in \attlist\}, S, \attlist, \attspace, \lambda, b$\Comment{Strength of prior ($\lambda$), budget (b)}
\State $t^*, N^* \leftarrow \operatorname{Calibrate}(\{\attpred_k\}, \dseed, U)$ \Comment{Calibration training of Alg:~\ref{alg:calibrate}}
\State $P(\attarm \mid \vx; t^*, N^*)$ is defined in Equation~\eqref{eqn:factor:assume} 
\State $\kappa = \frac{1}{|\dseed|}\sum_{(x,y,a)\in \dseed}\text{Agree}(S(x), y)$ \Comment{Prior accuracy}
\State c0, c1 $= \lambda\mathbbm{1}_{\abs{\attspace}}(1-\kappa), \lambda\mathbbm{1}_{\abs{\attspace}}\kappa$ \Comment{Initialize observation accumulators}
\State E $= \emptyset$ \Comment{Set of explored examples}
\For{$(\vx, y, \va) \in \dseed$}\Comment{Warm start with $\dseed$}
\State $c0[\va] = c0[\va] + (1-\match(S(\vx), y))$
\State $c1[\va] = c1[\va] + \match(S(x), y)$
\EndFor
\State Initialize $\rho(\va)$ with two GPs as described in Section~\ref{sec:betagp}, Equation~\eqref{eq:model}
\State Fit $\rho(\va)$ on c0, c1 as shown in~\eqref{eq:pointLLSimple},~\eqref{eqn:elbo}.
\While{$|E| < b$:}
\State $\hat\attarm = \argmax_{\attarm\in \attspace}\mathbb{V}[\rho(\attarm)]$ \Comment{Pick the arm with highest variance}
\State $\hat\vx = \argmax_{\{x\in U, x\not\in \text{E}\}} P(\hat\attarm\mid \vx)$ \Comment{Unexplored arm with highest affiliation}
\State Add $\hat{x}$ to E
\State Obtain the true label of $\hat x$: $\hat y$
\State c = \match($S(\hat x), \hat y$)
\For{$\attarm \in \attspace$}
\State c0[$\attarm$] = c0[$\attarm$] + (1-c)P($\hat\attarm\mid \hat x$)
\State c1[$\attarm$] = c1[$\attarm$] + cP($\hat\attarm\mid \vx$)
\EndFor
\State Fit again as in Line 11
\EndWhile
\State \Return $\rho$
\end{algorithmic}
\label{alg:main}
\end{algorithm}

\begin{algorithm}
\caption{Attribute Model Calibration (Section~\ref{sec:aaa:AttribNoise})}
\begin{algorithmic}[1]
\Require $\dseed , U, \{\attpred_k, k\in \attlist\}, \eta$
\State Initialize $t, N$
\State converged = False, $\tau$ = $10^{-3}$
\While{not converged}
\State d, u$^\prime$ $\leftarrow$ batch(D), batch(U) \Comment{sample a subset for batch processing}
\State u = \{($\vx$, \{$M_k(\vx)\mid k\in A$\}) for $\vx$ in u$^\prime$\}
\State LL = Eqn~\eqref{eqn:max_likelihood} on d, u
\State t, N = optimizer-update($\eta, \nabla_t LL, \nabla_N LL$)
\State converged = True if LL < $\tau$
\EndWhile
\State \Return $t, N$
\end{algorithmic}
\label{alg:calibrate}
\end{algorithm}

\end{document}